\pdfoutput=1
\documentclass[sigconf]{acmart}

\AtBeginDocument{%
  \providecommand\BibTeX{{%
    \normalfont B\kern-0.5em{\scshape i\kern-0.25em b}\kern-0.8em\TeX}}}

\renewcommand\footnotetextcopyrightpermission[1]{}

\setcopyright{none} 

\usepackage{amsfonts,amsmath}
\usepackage{mathtools}
\usepackage{xcolor}
\usepackage{enumitem}
\usepackage{subcaption}
\usepackage{bm, bbm}
\usepackage{xspace}
\usepackage{color, colortbl}
\usepackage{tabularray}
\usepackage{pifont}
\usepackage{arydshln}
\usepackage{romannum}
\usepackage{graphicx}
\usepackage{tikz}

\definecolor{CRed}{RGB}{255 204 204}
\definecolor{CGreen}{RGB}{102 255 102}
\definecolor{CBlue}{RGB}{153 204 255}

\usepackage{algorithm}
\usepackage{algpseudocode}
\algrenewcommand\algorithmicrequire{\textbf{Input:}}
\algrenewcommand\algorithmicensure{\textbf{Output:}}

\newcommand{\fingerprint}{FLARE\xspace}
\newcommand{\vect}[1]{\boldsymbol{#1}}
\DeclareMathOperator*{\argmax}{argmax}

\newcommand{\cmark}{\ding{51}}%
\newcommand{\xmark}{\ding{55}}%

\newcommand{\adversary}{$\mathcal{A}$\xspace}

\newcommand{\victim}{$\mathcal{V}$\xspace}

\newcommand{\verifier}{$\mathcal{J}$\xspace}

\newcommand{\redpen}[1]{}

\newcommand{\newtext}[1]{{#1}}

\begin{document}

\title{FLARE: Fingerprinting Deep Reinforcement Learning Agents using Universal Adversarial Masks}

\author{Buse G.  A.  Tekgul}
\affiliation{%
\institution{Nokia Bell Labs \& Aalto University}
\city{Espoo}
\country{Finland}}
\email{buse.atli_tekgul@nokia-bell-labs.com}

\author{N. Asokan}
\affiliation{%
\institution{University of Waterloo \& Aalto University}
\city{Waterloo}
\country{Canada}}
\email{asokan@acm.org}

\renewcommand{\shortauthors}{Tekgul and Asokan}

\begin{abstract}
We propose \fingerprint, the first fingerprinting mechanism to verify whether a suspected Deep Reinforcement Learning (DRL) policy is an illegitimate copy of another (victim) policy. We first show that it is possible to find non-transferable, universal adversarial masks, i.e., perturbations, to generate adversarial examples that can successfully transfer from a victim policy to its modified versions but not to independently trained policies. 
\fingerprint employs these masks as fingerprints to verify the true ownership of stolen DRL policies by measuring an action agreement value over states perturbed by such masks. Our empirical evaluations show that \fingerprint is effective (100\% action agreement on stolen copies) and does not falsely accuse independent policies (no false positives). \fingerprint is also robust to model modification attacks and cannot be easily evaded by more informed adversaries without negatively impacting agent performance. We also show that not all universal adversarial masks are suitable candidates for fingerprints due to the inherent characteristics of DRL policies. The spatio-temporal dynamics of DRL problems and sequential decision-making process make characterizing the decision boundary of DRL policies more difficult, as well as searching for universal masks that capture the geometry of it. 
\end{abstract}

%
%


%



\maketitle
\pagestyle{plain}

\section{Introduction}
\label{sec:Introduction}

Deep reinforcement learning (DRL) has emerged as a promising technique for building intelligent agents due to its ability to learn from and interact with high-dimensional input data.
Following the work of Mnih et al.~\cite{mnih2015human}, which shows that DRL has exceeded human-level performance in Atari games, it has been successfully used in many real-world applications, including green data centers~\cite{li2019transformingcooling}, autonomous driving~\cite{kiran2021deep} and robotic manipulation~\cite{pmlr-v87-kalashnikov18a}. 

The commercial success and continuous improvement of DRL methods attract adversaries, leading them to look for and exploit vulnerabilities in DRL agents. DRL agents leverage the power of deep neural networks (DNNs) to improve agents' decision-making strategy, i.e.,\emph{ policy}. Therefore, known vulnerabilities of DNNs might also be valid for agents' policies. For example, considerable research has been devoted to evasion attacks against DRL policies using adversarial examples~\cite{huang2017adversarial,gleave2019adversarial,Inkawhich2020snooping,pan2022characterizing,tekgul2022real}, which are generally computed for input states that are representations of the environment received by the agent. 
Unlike evasion attacks and related defenses~\cite{lin2017detecting,zhang2020robust,oikarinen2021robust}, few studies investigated the ownership of DRL models~\cite{behzadan2019sequential,chen2021temporal,lounici2021yes} against model piracy attacks. The high training costs of DRLs and their business advantage lead adversaries to steal models and redistribute them unauthorized ways. To deter adversaries, it is crucial to have the technical means to identify the true ownership of illegitimate copies of DRL agents. 


Recently, DNN fingerprinting has been proposed as an ownership verification method~\cite{lukas2021deep, peng2022fingerprinting}. DNN fingerprinting aims to identify the inherent properties of the victim (original) model and use this information during verification. Current DNN fingerprinting methods leverage \emph{conferrable adversarial examples} (CAE)~\cite{lukas2021deep} or \emph{universal adversarial perturbations} (UAP fingerprinting)~\cite{peng2022fingerprinting}, as adversarial examples can characterize the DNN decision boundary. Conferrable adversarial examples~\cite{lukas2021deep} are a subclass of transferable adversarial examples that can successfully force the victim DNN model and its modified versions to make the same wrong predictions, but are not transferable to other independently trained models. Unlike CAE, UAP~\cite{peng2022fingerprinting} obtains universal adversarial perturbations from both victim and suspected models, and produces a similarity score using contrastive learning. Both methods are shown to be effective and robust ownership verification approaches, but adapting them in DRL has challenges. First, both methods query suspected DNN models with adversarial version of input samples with different labels that are selected from the training set. However, this is not possible in DRL due to dynamic environments and continuous agent-environment interaction. Therefore, these methods require constructing a special verification setup that has unconventional environment dynamics and completely changes the trajectory of the agent regardless of the task. Second, there is no one-to-one mapping between the input states and the corresponding actions in DRL. This makes fingerprint generation challenging, since there is no single optimal action for any clean state and no desired incorrect action for any adversarial state either.


In this paper, we propose \fingerprint, \emph{the first DRL fingerprinting scheme} designed for discrete reinforcement learning tasks and combines the idea behind CAE and UAP. 
The effectiveness of adversarial perturbations decreases as they transfer from one DRL policy or algorithm to another~\cite{huang2017adversarial}. This implies that it would be possible to find adversarial examples that are not transferable across DRL agents. However, using individual non-transferable adversarial examples for ownership verification might be impractical due to the problems mentioned above. Therefore, \fingerprint aims to generate \emph{non-transferable universal masks} as fingerprints, which are independent of input states, source actions, or target actions. Fingerprints computed by \fingerprint are instances of weaknesses that are inherent in the victim's DRL policy. \fingerprint leverages these weak points to verify the true ownership of \emph{suspected} policies. During verification, suspected agents receive states modified by applying a universal mask from the victim's fingerprint in a small time window while trying to complete their task. \fingerprint verifies the true ownership if the similarity between the actions of the suspected agent and the victim agent in the same fingerprinted states is greater than a threshold value. 
\fingerprint does not change the training procedure, and verification can be implemented at any time during deployment. 
Our main contributions are as follows:
\begin{enumerate}
    \item We propose \fingerprint, the first fingerprinting method to verify the ownership of DRL agents used in discrete tasks by leveraging non-transferable universal adversarial masks (Section~\ref{sec:Methodology}). We show that \fingerprint is an effective ownership verification method with no false positives (Section~\ref{ssec:reliability}).\footnote{The code to reproduce our experiments is available on 
    \url{https://github.com/ssg-research/FLARE}\label{footnoteref}.}
    \item We verify the robustness of \fingerprint against model modification attacks (e.g., fine-tuning and pruning) on 6 different DRL agents trained using two different games of the Arcade Learning Environment~\cite{bellemare2013arcade}. We also show that well-informed adversaries cannot easily evade verification without sacrificing agent performance, and \fingerprint is robust against false claims made by malicious accusers. (Section~\ref{ssec:robustness}).
    \item We empirically demonstrate that universal adversarial perturbations generated by minimum-distance methods~\cite{moosavi2017universal,peng2022fingerprinting} are not good candidates for DRL fingerprinting. These perturbations are not unique weaknesses of DRL policies by design and fail against model modification attacks (Section~\ref{sec:Discussion}).
\end{enumerate}
\section{Background}
\label{sec:Background}
\subsection{Deep Reinforcement Learning}
\subsubsection{Reinforcement Learning} A typical reinforcement learning (RL) problem is modeled as a 5-tuple Markov Decision Process (MDP) $(S,A,P,R,\gamma)$, where $S$ denotes the state space, $A$ is the action space, $P$ symbolizes the state transition probability (i.e., environment dynamics), $R$ is the reward function, and $\gamma \in [0,1]$ denotes the discount factor used to calculate the discounted cumulative reward, i.e., \emph{return}. In this setting, the RL agent receives a state $\vect{s}_t \in S$ at the time step $t$, performs an action $a_t \in A$, and then subsequently receives a reward $r_{t+1}$ as well as the next state $\vect{s}_{t+1}$ based on $P(\vect{s}_{t+1}|\vect{s}_t,a_t)$. The objective of an RL agent is to maximize its expected return by interacting with the environment and to obtain an optimal policy $\pi(a | \vect{s}): S \rightarrow A$ that outputs an optimal action (the action that gives the maximum expected return over all actions) for any given state. During training, the policy is optimized recursively by calculating the expected return over states using the Bellman equation~\cite{SuttonBarto1998}. In this work, we consider states to be fully observable and finite-horizon tasks (i.e., an episode is completed when a stopping criterion is reached). Therefore, the discounted return at a time step $t$ is calculated as $R_{t} = \sum_{k=t}^{T}\gamma^{k-t}r_k$ where T is the final time step in a single episode. We also focus on tasks with a discrete action space, where one-hot vectors can be used to distinguish one action from every other action.

\subsubsection{Deep Reinforcement Learning (DRL)} When the state space $S$ is too complex and high-dimensional, deep neural networks (DNNs) can be useful to approximate policy $\pi(a | \vect{s})$. In this work, we assume that the environment is dynamic, as in real-world applications. Model-free DRL methods are the preferred approach in this setting, since these methods do not require estimating the dynamics of the environment. Two typical model-free DRL methods approximate $\pi$: value-based and policy-based methods. Value-based~\cite{mnih2015human} methods approximate the action value function $Q^{\pi}(\vect{s},a)$ which computes the estimated return of state $\vect{s}_t$ if the agent chooses the action $a_{t}$ and then follows the current policy. The optimal policy is implicitly obtained once $Q^{\pi}$ is optimized. Policy-based methods~\cite{mnih2016asynchronous,schulman2017proximal} first parameterize the policy $\pi(a|\vect{s}, \theta)$ and then optimize it by updating the parameters $\theta$ through the gradient ascent. 

In this paper, we use $\pi$ to symbolize the optimal policy obtained during training and $\hat{\pi}$ to denote the optimal action $a_t$ decided by $\pi$ for the input state $\vect{s}_t$, where $a_t = \hat{\pi}(\vect{s}_t)$. 

\subsection{Adversarial Examples}
\label{ssec:adversarial}

\subsubsection{Adversarial Examples in DNN} An adversarial example $\vect{x}'$ is an intentionally modified input sample $\vect{x} \in X$ with an imperceptible amount of noise $\vect{r}$ to force a DNN model $f: X \rightarrow Y$ into producing incorrect predictions $\hat{f}$. Targeted adversarial examples are labeled with $y'$, the intended (incorrect) prediction, in advance to satisfy $y' = \hat{f}(\vect{x}')$ and $ y' \neq \hat{f}(\vect{x})$, while untargeted adversarial examples aim to evade the correct prediction, i.e., $\hat{f}(\vect{x}') \neq \hat{f}(\vect{x})$. Untargeted adversarial examples against a victim DNN model $f$ are computed by solving an optimization problem, 
\begin{gather}\label{eq:advexamples}
      \argmax_{\vect{x}'}\mathcal{L}(f(\vect{x}'),\hat{f}(\vect{x}))\;
    \text{\small s.t.: } \lVert{\vect{x}'- \vect{x}}\rVert_{p} = \lVert{\vect{r}}\rVert_p \leq \epsilon,
\end{gather}
where $\mathcal{L}$ denotes the prediction loss of $f$. This formulation is used by \emph{maximum-confidence} adversarial example generation methods~\cite{demontis2019adversarial} that maximize $\mathcal{L}$ while constraining the amount of perturbation with $\epsilon$. On the contrary, the \emph{minimum-distance} methods aim to minimize the sufficient amount of perturbation that changes the prediction~\cite{moosavi2017universal}. 

An adversarial example $\vect{x}'$ calculated against one model $f$ and successfully misleads it can \emph{transfer} across other models, i.e., fools $f^{*}$ that are trained for the same task. The transferability of an adversarial example increases when the source model $f$ and the target models $f^{*}$ learn similar decision boundaries~\cite{tramer2017space}. Since maximum-confidence adversarial examples are misclassified with higher confidence, they have a higher transferability rate than minimum-confidence adversarial examples~\cite{demontis2019adversarial}.

The definition of an adversarial example in DRL differs according to the target component of the victim agent and the overall goal~\cite{huang2017adversarial,gleave2019adversarial, weng2020toward,MankowitzL2020Robust,tekgul2022real,pan2022characterizing}.
In this work, we consider adversarial states $\vect{s}'$ that mislead the policy $\pi$: $\hat{\pi}(\vect{s}') \neq \hat{\pi}(\vect{s})$, $\lVert{\vect{s}'-\vect{s}}\rVert_{p} = \lVert{\vect{r}}\rVert_p$, and set the norm $p$ to $\infty$.

\subsubsection{Universal Adversarial Perturbations} 
Instead of computing individual adversarial examples, Moosavi et al.~\cite{moosavi2017universal} propose finding a perturbation vector $\vect{r}$ that fools the DNN model $\hat{f}(\vect{x} + \vect{r}) \neq \hat{f}(\vect{x})$ on almost all data points $\vect{x}$ sampled from the same distribution as the dataset $\mathcal{D}_{train}$ used for training $f$. The optimization problem in Equation~\ref{eq:advexamples} is modified to find universal perturbations as 

\begin{gather}\label{eq:foolingrate}
\mathbb{P}_{\vect{x} \sim \mu}(\hat{f}(\vect{x} + \vect{r}) \neq \hat{f}(\vect{x})) \geq \delta_{\vect{r}}\;
\text{\small s.t.: } \lVert{\vect{r}}\rVert_p \leq \epsilon,
\end{gather}
where $\delta_{\vect{r}}$ denotes the desired \emph{fooling rate} of $\vect{r}$ for all $\vect{x}$ sampled from a dataset $D$ with distribution $\mu$.

Following Moosavi et al.'s initial work~\cite{moosavi2017universal}, several different techniques are proposed to generate universal adversarial perturbations. For example, Mopuri et al.~\cite{mopuri2018NAG} train a generative adversarial network to model the distribution of universal adversarial perturbations for a target DNN classification model and produce diverse perturbations that achieve a high $\delta_{\vect{r}}$. Liu et al.~\cite{Liu2019Universal} generate a universal adversarial perturbation that does not require training data and exploits the uncertainty of the model at each DNN layer. 


\subsection{Ownership Verification via Fingerprinting}
Ownership verification in machine learning (ML) refers to a type of defense against model theft and extraction attacks by deterrence. Model owners can reduce the incentive for such attacks by identifying and verifying the true ownership of stolen models. DNN model fingerprinting is a well-known ownership verification technique. DNN fingerprinting methods identify unique knowledge that characterizes the victim model (fingerprint generation) and later use this information to verify whether the suspected model is derived from the victim model (fingerprint verification). For example, Ciao et al.~\cite{Cao2021IPGuard} use adversarial example generation methods to extract data points near the decision boundary of DNN classifiers, label them as fingerprints, and utilize them along with their labels to detect piracy models. Lukas et al.~\cite{lukas2021deep} fingerprint DNN models through conferrable adversarial examples (CAE) that can successfully transfer from the source model to its modified versions, but not to other DNN models independently trained for the same classification task. To verify the fingerprint in a suspected model, CAE measures the error rate between the predictions of victim and suspected models, and the verdict is delivered based on a decision threshold. CAEs employ predictions of different independently trained models and modified versions of the victim model to compute fingerprints. Therefore, CAE has a high computational cost, since it requires training multiple modified and independent models to extract conferrable adversarial examples. Peng et al.~\cite{peng2022fingerprinting} propose using universal adversarial perturbations (UAP) as fingerprints. During verification, previously computed UAPs for both victim and suspected models are mapped to a joint representation space, and contrastive learning is used to measure a similarity score in this projected space.

Adopting both UAP and CAE in DRL settings faces similar challenges. First, the verification episodes should include adversarial states that are completely different from each other, and also from a normal test episode during deployment. \newtext{Second, in CAE, the predictions of multiple models having a good performance might be close to each other for the same input samples, since these models are trained over the same labeled dataset.}
However, there is no single predefined optimal action for input states in DRL. When agents receive the same state, they might act differently to perform the task due to their unique and different policies. 
Third, UAP fingerprinting uniformly selects data samples that are from different source classes and moves them towards different target classes in DNNs, but there is no one-to-one mapping between input states and corresponding optimal actions to obtain useful fingerprints in DRL settings.

\section{Methodology}
\label{sec:Methodology}
\subsection{Adversary Model}
\label{ssec:adversarymodel}
The adversary \adversary's goal is to obtain an illegal copy of the victim agent's (\victim) policy $\pi_{\mathcal{V}}$ without being detected. \adversary has economic incentives and aims to illegally monetize stolen policy $\pi_{\mathcal{A}}$ using a surrogate DRL agent. $\pi_{\mathcal{V}}$ can be leaked by exploiting hardware/software vulnerabilities~\cite{yan2020cache} of different components within \victim. Furthermore, \adversary seeks to prevent traceback. Therefore, \adversary attempts to degrade the effectiveness of possible ownership verification methods by modifying $\pi_{\mathcal{A}}$, without incurring any substantial drop in $\pi_{\mathcal{A}}$'s return. 

\subsubsection{Adversary's capabilities.}\adversary has computational capabilities and access to the similar environment that $\pi_{\mathcal{V}}$ was trained on, but it cannot reproduce the same training episodes. 
One can argue that \adversary can also train its own policy, but we assume that it cannot obtain a policy as good as $\pi_{\mathcal{V}}$ due to nondeterminism (e.g., network architecture, difference in environment dynamics, DRL algorithm, hyperparameter selection, difference in computational resources, etc.). \adversary presumes that there might be an ownership verification mechanism, but does not know the exact algorithm. Based on this assumption, we also consider the existence of \emph{well-informed} adversaries\footnote{\newtext{ML literature commonly uses the term ``adaptive'' to refer to adversaries who are aware of deployed defenses. In security literature, it is customary to assume that \emph{all} adversaries are aware of the defenses, and the term ``adaptive'' is used for adversaries who are able to \emph{dynamically modify} their attack strategy based on what they learn about the defenses \emph{during} the attack. We use the term ``well-informed'' to refer to such adversaries so that our usage does not conflict with either ML or security literature.}} knowing that ownership verification is performed by fingerprinting and adversarial examples. If well-informed \adversary knows the complete procedure of the fingerprinting process, then it can forge its own fingerprints to create ambiguity in verification. However, this could be prevented with \fingerprint if $\pi_{\mathcal{V}}$ and the corresponding fingerprints are securely time-stamped and registered in a bulletin or provided to a trusted third party~\cite{szyller2021dawn}.

\subsubsection{Verifier's Capabilities.} A verifier (judge, \verifier) is a trusted third party independent of both \victim and \adversary. Given a suspected DRL agent $\mathcal{S} $ with policy $\pi_{\mathcal{S}}$ and fingerprints provided by \victim, the duty of \verifier is to determine whether $\pi_{\mathcal{S}}$ can be traced back to $\pi_{\mathcal{V}}$ and demonstrate the true ownership. \verifier has black-box access to $\mathcal{S}$, i.e., it does not know the algorithm and parameters of $\pi_{\mathcal{S}}$. \verifier can modify the environment without introducing any temporal latency or suspending the task. If the verification uses time stamps, it provides anteriority to \verifier to resolve any ambiguity. We also give \verifier computational capabilities to train and search for independent policies used for the same task if there is a need to validate that fingerprints are unique to the original model and do not transfer to independent models. We also define that a good fingerprinting mechanism should satisfy the following requirements: 
\begin{enumerate}[leftmargin=*]
\item \textbf{Effectiveness}: Successful ownership verification of stolen policies, \newtext{i.e., maximizing true positives.}
\item \textbf{Integrity}: Avoiding accidental accusations of independently trained policies, \newtext{i.e., minimizing false positives}.
\item \textbf{Robustness}: Withstanding model modification and evasion attacks. This is achieved if either the ownership of the modified policy is still successfully verified or the modification results in a substantial decrease in utility measured by the agent performance. 
\end{enumerate}
Fingerprinting algorithms do not necessarily aim for \emph{utility} (i.e., maintaining the quality of the suspected model on fingerprints), as they typically use adversarial examples during verification~\cite{lukas2021deep,peng2022fingerprinting} and the desired outcomes for fingerprints contain incorrect predictions. Therefore, we did not include utility as a requirement. However, we still restrict \fingerprint based on the utility concept, so that agents can still maintain their overall performance and complete the task without a significant performance degradation in episodes that include the verification phase. 

\subsection{Universal Adversarial Masks as Fingerprints}
\label{ssec:fingerprint-design}

\fingerprint aims to find a set of adversarial masks that can fool the original agent in any input state to which it is added, but cannot transfer to independently trained agents. Lukas et al.~\cite{lukas2021deep} define a similar property for classifiers called ``conferrability''. Conferrable adversarial examples can transfer from the original classifier to its derivatives but not to independently trained classifiers. In contrast, \fingerprint does not generate individual adversarial examples but instead searches for universal adversarial masks that can be used to generate conferrable adversarial examples.

\subsubsection{Fingerprint Generation.} 

During fingerprint generation, \fingerprint first computes the universal adversarial mask using the original policy $\pi_{\mathcal{V}}$ and independently trained models $\pi_{i}, (i \in \mathcal{I})$ that have the same DNN architecture. \fingerprint aims to find a universal mask $\vect{r}$ that maximizes the loss function in Equation~\ref{eq:lossfunction} and is bounded by $\epsilon$ in $l_{\infty}$-norm.
\begin{equation}\label{eq:lossfunction}
    \mathcal{L}(\pi_{\mathcal{V}}(\vect{s}_t + \vect{r}), \hat{\pi}_{\mathcal{V}}(\vect{s}_t)) - \mathbbm{1}_{(\hat{\pi}_\mathcal{V}(\vect{s}_t) = \hat{\pi}_{i}(\vect{s}_t))}
    \mathcal{L}(\pi_{i}(\vect{s}_t + \vect{r}), \hat{\pi}_{i}(\vect{s}_t))
\end{equation}

The first part of Equation~\ref{eq:lossfunction} maximizes the categorical cross-entropy loss between $\pi_{\mathcal{V}}$'s predictions for clean and adversarial states \newtext{using the log-probability vector for all actions $\pi_{\mathcal{V}}$ in adversarial state and the performed action $\hat{\pi}_{\mathcal{V}}$ in the clean version of that state. The second part minimizes the categorical cross-entropy loss between adversarial states $\vect{s}_{t} + \vect{r}$ computed for $\pi_i$ and their clean counterparts $\vect{s}_t$ only if the predicted action for $\vect{s}_t$ is the same for both $\pi_{\mathcal{V}}$ and $\pi_{i}$. The modified loss function ensures that the same $\vect{s}_t + \vect{r}$ cannot mislead $\pi_{\mathcal{V}}$ and $\pi_{i}$ in the same way, even if $\hat{\pi}_{i}$ produces a suboptimal action.} \newtext{\fingerprint uses untargeted adversarial examples as fingerprints (see Section~\ref{ssec:adversarial}), so the solution of Equation~\ref{eq:lossfunction} forces $\pi_{\mathcal{V}}$ into the incorrect action in $\vect{s}_{t} + \vect{r}$, but has a minimum effect on $\pi_i$.} Multiple independently trained policies are used to calculate the second part of Equation~\ref{eq:lossfunction} for each $i \in \mathcal{I}$ by taking the average of individual losses. A universal adversarial mask should also achieve a high fooling rate $\delta_{\vect{r}}$ as presented in Equation~\ref{eq:foolingrate}.


\begin{algorithm}[t]
\caption{Fingerprint generation}
\label{alg:fingerprint_generation}
\begin{algorithmic}[1]
\Require{$\mathcal{D}_{flare}$: Fingerprint generation set}
\Ensure{FRL: Fingerprint list}
\State \text{parameters:} $\tau_{nts}, \tau_{\delta}, 
n_{\text{episodes}}, n_{\text{FRL}}$
\State $\text{FRL} = [\;]$.
\For{$eps \leq$ $n_{\text{episodes}}$}
    \State \text{Generate} $\vect{r}_{candidate}$ \text{from}  $\mathcal{D}_{flare}$
    \State Compute $nts(\vect{r}_{candidate})$ using $\mathcal{I}$ and $\forall \vect{s}_t \in eps$
    \If {$nts \geq \tau_{nts} \;\textbf{and}\; \delta_{\vect{r}_{candidate}} \geq \tau_{\delta}$}
        \State Add $\vect{r}_{candidate}$ into FRL
    \EndIf
    \If {$len(\text{FRL}) == n_{\text{FRL}}$}
        \State return FRL
    \EndIf
\EndFor
\State \textbf{return} \text{FRL}
\end{algorithmic}
\end{algorithm}

To ensure universality, \fingerprint uses an approach similar to~\cite{pan2022characterizing} when solving Equation~\ref{eq:lossfunction}. First, \victim completes one episode and the observed states are saved in a training set $\mathcal{D}_{flare}$. Then \fingerprint computes the average gradient of the loss function in Equation~\ref{eq:lossfunction} w.r.t. $k$ states randomly sampled from $\mathcal{D}_{flare}$. This enables \fingerprint to generate $len(\mathcal{D}_{flare})\choose{k}$ different universal adversarial masks as a fingerprint candidate. After generating the fingerprint candidate, \fingerprint checks its \emph{non-transferability score}. We compute the non-transferability score $(nts)$ for a universal adversarial mask $\vect{r}$ on an episode $eps$ (that $\pi_{\mathcal{V}}$ follows) as 
\begin{equation}\label{eq:nts}
    nts(\vect{r}, eps) = \delta_{\vect{r}, eps}\times max_{i \in \mathcal{I}}(1-AA(\pi_{\mathcal{V}}, \pi_{i}, \vect{s}, \vect{r})),
\end{equation}
where $\delta_{\vect{r}, eps}$ refers to the fooling rate measured for $\pi_{\mathcal{V}}$ using all $\vect{s}$ observed in $eps$. $AA$ denotes \emph{action agreement} and is calculated as
\begin{equation}
    AA(\pi_i, \pi_j, \vect{s}, \vect{r}) = \frac{1}{N}\sum_{\substack{t=0}}^{t=N} 
    \mathbbm{1}_{(\hat{\pi}_i(\vect{s}_t + \vect{r}) = \hat{\pi}_j(\vect{s}_t + \vect{r}))},
\end{equation} 
where $N$ refers to the length of one full episode $eps$ that $\pi_{\mathcal{V}}$ follows.

\fingerprint only accepts the candidate $\vect{r}_{candidate}$ as a valid fingerprint if $nts(\vect{r}_{candidate})$ is greater than a threshold value $\tau_{nts}$ and achieves a fooling rate $\delta_{\vect{r}_{candidate}}$ higher than $\tau_{\delta}$ over a single $eps$. How \fingerprint decides whether to include a universal adversarial mask in a fingerprint list FLR is presented in Algorithm \ref{alg:fingerprint_generation}.

\subsubsection{Fingerprint Verification.} For fingerprint verification, the verifier \verifier has given a fingerprint set FLR. \verifier first observes the interactions between the suspected agent $\mathcal{S}$ and the environment to estimate the total number of states $N$ that occur during a single episode. Then, for each subsequent episode, \verifier adds one fingerprint starting from a random state at time $t_{start}$ over a short time window of length $M$ to preserve the return in an acceptable range. \victim is also queried with the adversarial states $\vect{s}_t + \vect{r}$ that the suspected agent receives. For each fingerprint, $AA$ is calculated as $1/M\sum_{\substack{t=t_{start}}}^{t_{start}+M-1} AA(\pi_{\mathcal{V}}, \pi_{\mathcal{S}}, \vect{s}_t, \vect{r})$. If $AA$ for a single fingerprint exceeds a decision threshold $AA \geq 0.5$, that fingerprint produces supporting evidence to verify that the suspected model is the stolen copy. The final verdict (stolen vs. independent) is made based on the \emph{majority vote}. \fingerprint also returns $AA$ averaged on all fingerprints to quantify the confidence in the final decision. The verification procedure is summarized by Algorithm~\ref{alg:fingerprint_verification}.

\begin{algorithm}[t]
\caption{Fingerprint verification}
\label{alg:fingerprint_verification}
\begin{algorithmic}[1]
\Require{FRL, $\pi_{\mathcal{V}}$, $\pi_{\mathcal{S}}$: Fingerprint list, victim and suspected policies}
\Ensure{$AA$, $Mvote$: action agreement, majority vote}
\State $AA = [\;], Mvote=0, Tvote=0.$
\State{Run a single episode with $\pi_{\mathcal{S}}$, save total number of states $N$}
\For{$i,\vect{r}_i\; \text{in}\; (range(\text{FRL}), \text{FRL})$}
    \State{$AA_i = 0.0$}
    \State {Generate random $t_{start} \in [0,min(N, N-M)]$}
    \State {Run a test episode with $\pi_{\mathcal{S}}$}
    \While {test episode of $\pi_{\mathcal{S}}$ not finished}
    \State {Calculate $AA_i$ over time steps $t \in  [t_{start}, t_{start}+M)$}
    \EndWhile
    \State{$Mvote \,+= 1$ \textbf{if} $(AA_i \geq 0.5)$, $Tvote\,+=1$}
    \State{Add $AA_i$ into $AA$}
    \State{Decision: Stolen \textbf{if} $Mvote > (Tvote-Mvote)$}
\EndFor
\State \textbf{return} Decision, $Mvote$, mean and std of $AA$
\end{algorithmic}
\end{algorithm}

\section{Empirical Analysis}
\label{sec:Experiments}

\newcommand{\greenline}{\raisebox{0.0pt}{\tikz{\draw[-, green!40,solid,line width = 6pt](0,0) -- (5mm,0);}}}

\newcommand{\blueline}{\raisebox{0.0pt}{\tikz{\draw[-, blue!30,solid,line width = 6pt](0,0) -- (5mm,0);}}}

\newcommand{\yellowline}{\raisebox{0.0pt}{\tikz{\draw[-, yellow!40,solid,line width = 6pt](0,0) -- (5mm,0);}}}

\newcommand{\redline}{\raisebox{0.0pt}{\tikz{\draw[-, red!40,solid,line width = 6pt](0,0) -- (5mm,0);}}}

\subsection{Experimental Setup}

We evaluated \fingerprint using the Arcade Learning Environment (ALE)~\cite{bellemare2013arcade}. We selected two different games, Pong and MsPacman, from ALE to train agents with three different model-free DRL algorithms: A2C~\cite{mnih2016asynchronous}, DQN~\cite{mnih2015human}, and PPO~\cite{schulman2017proximal}. Pong is a two-player game in which agents are trained to win against the computer, while MsPacman is a single-player game with the goal of achieving the highest score without crashing into enemies. When constructing the state information, we applied the pre-processing methods proposed in~\cite{mnih2015human}. Furthermore, for each victim \victim, we independently trained five additional policies $\pi_{i}\, (i \in \mathcal{I})$ that have the same DNN architecture and the DRL algorithm as \victim, and used them during fingerprint generation. In total, we obtained six victim policies and thirty independent policies. In Pong, the victim and the independent policies win the game with the highest score (+21). In MsPacman, it was harder to achieve similar high scores since states are more complex than Pong and depend on the position of multiple enemies. In both games, the score is used to quantify the agent's return. Appendix~\ref{ssec:appendixA1} presents software/hardware requirements for reproduction, as well as the average performance of all agents. 

\newtext{During fingerprint generation in DQN, \fingerprint uses the DNN approximating Q value function. For other algorithms, \fingerprint selects the policy network (e.g., the actor network in A2C) to compute fingerprints.} We set the maximum number of fingerprints $len(FRL)$ at 10, and the window size $M$ at 40. The discussion on the choice of $len(FRL)$ and $M$ is included in Appendix~\ref{ssec:appendixA3}. Other hyperparameters used in fingerprint generation are also listed in Appendix~\ref{ssec:appendixA3}. In our experimental setup, we used different random initialization for episodes used in training, fingerprint generation, verification, estimation of agent performance, modification attacks, and evasion attacks to ensure randomness in dynamic (and uncontrollable) environments.

\subsection{Effectiveness and Integrity}\label{ssec:reliability}

\begin{figure}[t]
\centering
\includegraphics[width=1.0\columnwidth]{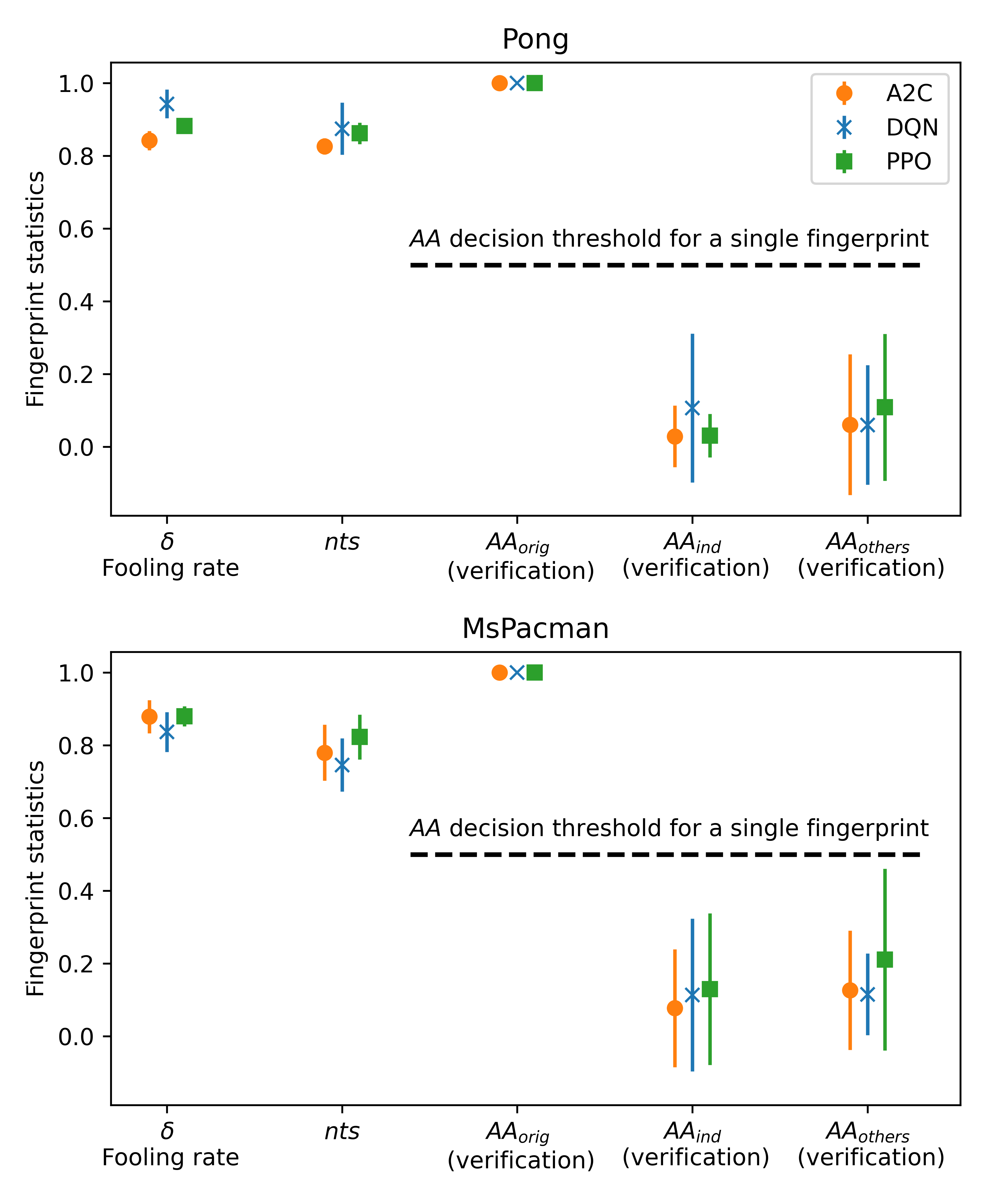} 
\caption{Various \fingerprint metrics averaged over 10 runs for all generated fingerprints. \fingerprint can successfully distinguish between the original model $AA_{orig}$ and independent models $AA_{ind}$, $AA_{others}$, while achieving high fooling rate $\delta$ and non-transferability score $nts$.}
\label{fig:fooling_rate}
\end{figure}

\begin{table*}[t]
\centering
\caption{Average impact, $AA$ and voting results (\cmark:Stolen, \xmark: Independent) for piracy policies that are 1) fine-tuned over a different number of episodes and 2) pruned and then fine-tuned over 200 episodes. $AA$ is averaged over 10 verification episodes, while impact is averaged over 10 test episodes. (\protect\greenline : Successful verification with $AA\geq 0.75$, \protect\blueline : Successful verification with $0.75 \geq AA \geq 0.50$, \protect\yellowline : Failed verification with high impact $\geq 0.4$, \protect\redline : Failed verification with low impact $ < 0.4$)}\label{tab:fine_tuning_pruning}
\begin{tblr}{colspec={ccccc|cccc},
cell{5}{3-7} = {green!40},
cell{5}{8} = {blue!30},
cell{5}{9} = {yellow!40},
cell{8}{3-6} = {green!40},
cell{8}{7} = {blue!30},
cell{8}{8-9} = {yellow!40},
cell{11}{3-6} = {green!40},
cell{11}{7-9} = {blue!30},
cell{11}{9} = {blue!30},
cell{14}{3-4} = {green!40},
cell{14}{5} = {blue!30},
cell{14}{6} = {blue!30},
cell{14}{7} = {blue!30},
cell{14}{8} = {blue!30},
cell{14}{9} = {blue!30},
cell{14}{3} = {green!40},
cell{17}{3-9} = {yellow!40},
cell{20}{3-9} = {yellow!40},}
\SetCell[r=2]{m,5em}\bf Game, DRL method & \SetCell[r=2]{b}\bf Stats &
\SetCell[c=3]{c}\bf Fine-tuning, \# of episodes & & & \SetCell[c=4]{c}\bf Pruning and fine-tuning, pruning levels (\%) & & & &\\ 
&  & \bf 50 & \bf 100 & \bf 200 & \bf 25 & \bf 50 & \bf 75 & \bf 90 \\
\hline
\SetCell[r=3]{m,4em}\bf Pong, A2C& Impact & $0.0 \pm 0.0$  & $0.0 \pm 0.0$ & $0.0 \pm 0.0$ 
& $0.0 \pm 0.0$  & $0.0 \pm 0.0$ & $0.0 \pm 0.0$  & $1.0 \pm 0.0$ \\
& $AA$ & $0.95 \pm 0.14$ & $0.95 \pm 0.14$& $0.94 \pm 0.10$ & $0.94 \pm 0.14$ & $0.91 \pm 0.25$& $0.67 \pm 0.42$ & $0.28 \pm 0.42$\\
& Votes & 10 \cmark / 0 \xmark  & 10 \cmark / 0 \xmark &  10 \cmark / 0 \xmark & 
10 \cmark / 0 \xmark & 9 \cmark / 1 \xmark & 6 \cmark / 4 \xmark  & 3 \cmark / 7 \xmark \\
\hline[dashed]
\SetCell[r=3]{m,4em} \bf Pong, DQN& Impact & $0.0 \pm 0.0$  & $0.0 \pm 0.0$ & $0.0 \pm 0.0$ 
& $0.0 \pm 0.0$  & $0.8 \pm 0.0$ & $1.0 \pm 0.0$  & $1.0 \pm 0.0$ \\
& $AA$ &  $0.94 \pm 0.05$ & $0.89\pm 0.14$ &  $0.90 \pm 0.17$ & $0.88 \pm 0.16$ & $0.66 \pm 0.38$& 
$0.09 \pm 0.17$& $0.27 \pm 0.4$\\
& Votes & 10 \cmark / 0 \xmark & 10 \cmark / 0 \xmark& 9 \cmark /1 \xmark & 
10 \cmark / 0 \xmark & 7 \cmark / 3 \xmark & 1 \cmark / 9 \xmark &3 \cmark / 7 \xmark\\
\hline[dashed]
\SetCell[r=3]{m,4em}\bf Pong, PPO& Impact & $0.0 \pm 0.0$  & $0.0 \pm 0.0$ & $0.0 \pm 0.0$ 
& $0.0 \pm 0.0$  & $0.0 \pm 0.0$ & $1.0 \pm 0.0$  & $1.0 \pm 0.0$ \\
& $AA$ & $0.88 \pm 0.23$ & $0.89\pm 0.25$ & $0.88\pm 0.30$ &  
$0.78 \pm 0.35$ &  $0.67 \pm 0.35$&  $0.65 \pm 0.41$ &  $0.71 \pm 0.39$ \\
& Votes & 9 \cmark /1 \xmark  &9 \cmark /1 \xmark  & 9 \cmark /1 \xmark  & 
7 \cmark / 3 \xmark &  7 \cmark / 3 \xmark  &  6 \cmark / 4 \xmark & 7 \cmark / 3 \xmark \\
\hline[dashed]
\SetCell[r=3]{m,5em}\bf MsPacman, A2C& Impact & $0.0 \pm 0.0$ & $0.0 \pm 0.0$ & $0.0 \pm 0.0$
& $0.39\pm0.19$ & $0.03\pm0.10$& $0.30\pm0.15$ & $0.73\pm0.11$\\
& $AA$ & $0.82 \pm 0.16$ & $0.75 \pm 0.29$ & $0.62 \pm 0.35$ &  
$0.71 \pm 0.28$  & $0.65 \pm 0.39$ &  $0.72 \pm 0.26$ & $0.59 \pm 0.23$  \\
& Votes & 9 \cmark /1 \xmark  & 8 \cmark /2 \xmark & 6 \cmark /4 \xmark & 
6 \cmark /4 \xmark & 7 \cmark /3 \xmark &  8 \cmark /2 \xmark &  6 \cmark /4 \xmark  \\
\hline[dashed]
\SetCell[r=3]{m,5em}\bf MsPacman, DQN& Impact & $0.79 \pm 0.11$ & $0.83 \pm 0.02$ & $0.87 \pm 0.03$
&  $0.79 \pm 0.11$ &  $0.74 \pm 0.09$ &  $0.86 \pm 0.01$ &  $0.71 \pm 0.43$ \\
& $AA$ & $0.23 \pm 0.34$ & $0.15 \pm 0.28$& $0.16 \pm 0.31$ & $0.38 \pm 0.44$ 
& $0.00 \pm 0.01$  & $0.59 \pm 0.46$ & $0.42 \pm 0.42$ \\
& Votes & 2 \cmark /8 \xmark & 1 \cmark /9 \xmark& 2 \cmark /8 \xmark & 4 \cmark /6 \xmark
& 0 \cmark /10 \xmark& 6 \cmark /4 \xmark & 4 \cmark /6 \xmark\\
\hline[dashed]
\SetCell[r=3]{m,5em} \bf MsPacman, PPO& Impact & $0.85 \pm 0.11$ & $0.40 \pm 0.26$ & $0.51 \pm 0.08$
& $0.52 \pm 0.15$ & $0.57 \pm 0.04$ &  $0.62 \pm 0.05$ & $0.66 \pm 0.19$\\
& $AA$ & $0.43 \pm 0.36$ & $0.11 \pm 0.16$ &  $0.25 \pm 0.32$& 
$0.26 \pm 0.36$ & $0.33 \pm 0.38$ &  $0.31 \pm 0.32$ & $0.13\pm 0.20$ \\
& Votes & 4 \cmark /6 \xmark & 0 \cmark /10 \xmark&  3 \cmark /7 \xmark& 
3 \cmark / 7 \xmark& 3 \cmark / 7 \xmark&  4 \cmark / 6 \xmark& 1 \cmark / 9 \xmark \\
\hline 
\end{tblr}
\end{table*}

Figure~\ref{fig:fooling_rate} summarizes various \fingerprint metrics (fooling rate $\delta$, non-transferability score $nts$ and action agreement $AA$ calculated on different policies) for three different DRL algorithms.
$AA_{orig}$ denotes $AA$ of the adversary's policy $\pi_{\mathcal{A}}$ which is identical to the victim policy $\pi_{\mathcal{V}}$. $AA_{ind}\,\text{(verification)}$ refers to $AA$ values of independent policies $\pi_{i}$ that share the same DRL algorithm as $\pi_{\mathcal{V}}$ and are used in Algorithm~\ref{alg:fingerprint_generation} (5 policies for each $\mathcal{V}$). We use the remaining 10 independent policies (having a different DRL algorithm from $\mathcal{V}$) trained for the same task to calculate $AA_{others}\,\text{(verification)}$. $AA_{orig}$ is much higher than the threshold value $0.5$, almost equal to $1.0$ in most cases. Furthermore, the average fooling rate of fingerprints is high, which proves that fingerprints successfully mislead $\pi_{\mathcal{A}}$. The average of $AA_{ind}\,\text{(verification)}$ and $AA_{others}\,\text{(verification)}$ are lower than $0.5$ in all cases, and the majority vote is always ```not stolen (independent)'' for any other $\pi_{i}$ that is not $\pi_{\mathcal{A}}$. \newtext{Results show that \fingerprint achieves a high detection rate while avoiding false accusations of independently trained ones.} Thus, we conclude that \fingerprint satisfies the effectiveness and integrity requirements. 

As shown in Figure~\ref{fig:fooling_rate}, $AA_{ind}$ and $AA_{others}$ show different variances for three DRL algorithms. We found that one or two fingerprints seldom produce $AA \geq 0.5$ for $\pi_{\mathcal{V}}$ and $\pi_{i}$, although they behave differently in the same clean states. This reveals that a single fingerprint rarely represents the same weakness of two separate policies, and the number of fingerprints should be high enough to satisfy integrity considering this phenomenon. 

\newtext{During verification, we set the threshold value to $0.5$ (a single fingerprint votes for ``stolen'' if $AA \geq 0.5$) over all experiments. However, it might be better to look at the full profile of the receiver operational characteristic (ROC) curves, which give a complete picture of the trade-off between false positive and true positive rates by varying the threshold value. We provide ROC curves for both Pong and MsPacman games in Appendix~\ref{ssec:appendixA4}.}

\textit{Utility} : As stated in Section~\ref{ssec:adversarymodel}, we do not consider utility a requirement for \fingerprint. However, based on the definition in~\cite{korkmaz2022deep}, we measure the \emph{impact} of the verification on the victim agent to ensure that it does not fail the task during verification. We measure the impact as:

\begin{equation}\label{eqn:impact}
    Impact = \frac{Return_{\mathcal{V}(test)} - Return_{\mathcal{V}(verification)}}{Return_{\mathcal{V}(test)} - Return_{\mathcal{V}_{min}(test)}}.
\end{equation}

$Return_{\mathcal{V}(test)}$ and $Return_{\mathcal{V}(verification)}$ are the average return of \victim in an episode initialized with the same start state (and with the same environment dynamics) with or without the verification. $Return_{\mathcal{V}_{min}(test)}$ is the return of $\mathcal{V}$ if it chooses the worst possible actions for each state in the same episode. The results presented in Appendix~\ref{ssec:appendixA5} show that the average impact on agent performance is $0.02$ and $0.22$ in MsPacman and Pong, respectively. We also found that the return never drops to $Return_{\mathcal{V}_{min}(test)}$ during verification. Thus, we conclude that agents continue their task without a significant impact after the verification phase ends.

\subsection{Robustness}\label{ssec:robustness}

\begin{figure*}[t]
\centering
\includegraphics[width=0.9\textwidth]{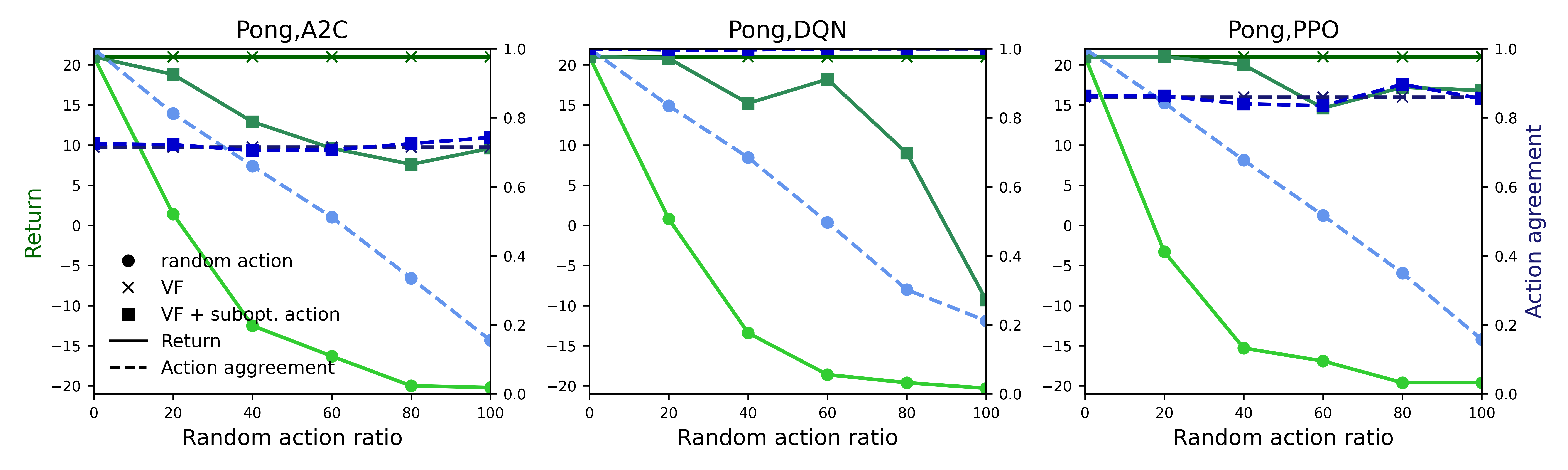}
\includegraphics[width=0.9\textwidth]{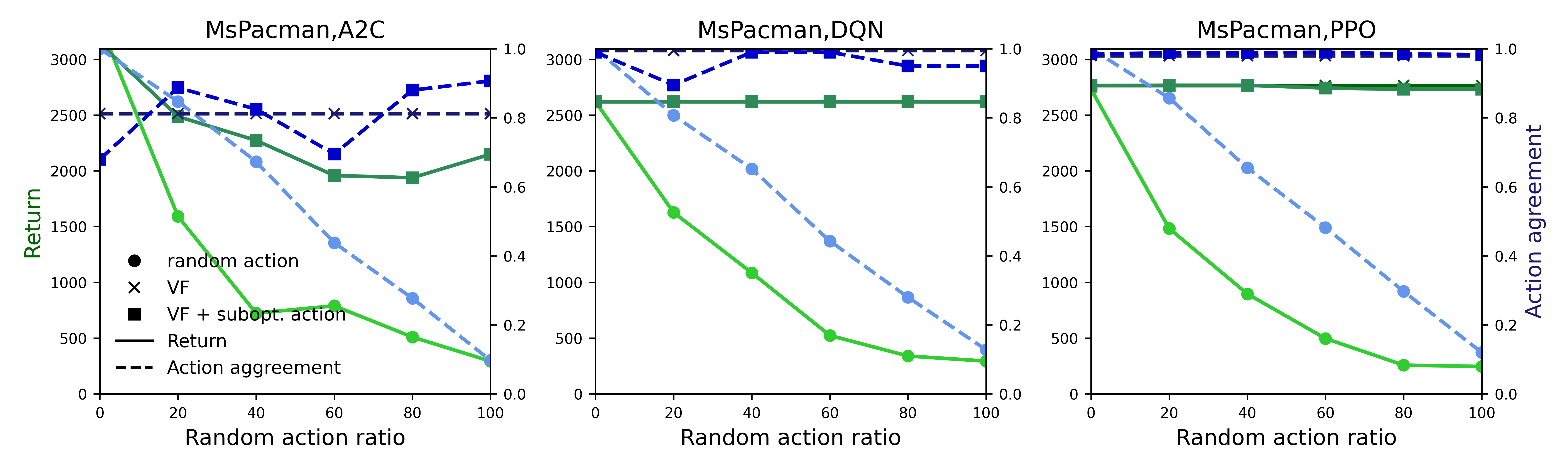}
\caption{$AA$ and return when attacker implements random action with different ratios, Visual Foresight (VF), and VF+suboptimal action as evasion against \fingerprint. $AA$ is averaged in 10 verification episodes, whereas the return is averaged in 10 test episodes. \newtext{Solid lines represent the return while dashed lines refer to $AA$. Each plot includes three solid and dashed lines (some of which overlap), and different markings on these lines refer to a specific evasion method.}}\label{fig:evasion_results}
\end{figure*} 

\subsubsection{Robustness Against Model Modification Attacks.} Adversary \adversary could modify the stolen policy $\pi_{\mathcal{A}}$ by carefully retraining it to preserve agent performance while trying to suppress evidence used for verification. We consider two common types of model modification attacks, fine-tuning~\cite{sharif2014cnn} and weight pruning~\cite{han2015learning}, where \adversary is aware of the existence of an ownership verification technique but does not know the type of it. We implemented fine-tuning by retraining $\pi_{\mathcal{A}}$ in an additional 200 episodes and decreasing the learning rate by $100$ to maintain agent performance. For pruning, we first performed global pruning, i.e., removed a percentage of the lowest connections across the DNN model. After pruning, we fine-tuned the pruned model over 200 episodes. 

To evaluate the robustness requirement, we computed the majority vote and $AA$ values of the stolen $\pi_{\mathcal{A}}$ and modified policies $\pi_{\mathcal{A}^*}$ on the verification episodes. We also measured the impact of the modification on model utility (agent performance) by changing Equation~\ref{eqn:impact} to:
\begin{equation}\label{eqn:impact2}
    Impact = \frac{Return_{\mathcal{A}(test)} - Return_{\mathcal{A}^*(test)}}{Return_{\mathcal{A}(test)} - Return_{A_{min}(test)}},
\end{equation}
where $Return_{\mathcal{A}^*}$ is the return of stolen and modified policy, and $Return_{\mathcal{A}}$ denotes the return of stolen (unmodified) policy over the same test episodes. Based on the results on the impact of verification on utility, we generously set the maximum allowable impact as $0.4$ for modification attacks, indicating that $Return_{\mathcal{A}^*(test)}$ would fall a little more than halfway between $Return_{\mathcal{A}(test)}$ and $Return_{\mathcal{A}_{min}(test)}$.

Table~\ref{tab:fine_tuning_pruning} shows the robustness evaluation of \fingerprint against model modification attacks. As shown in the table, \fingerprint successfully verifies fine-tuned Pong agents with high $AA$ values. \fingerprint usually results in a failed verification of fine-tuned MsPacman agents. However, the impact of modification is exceptionally high for these cases. A similar conclusion can be drawn from the pruning results. An increase in the pruning level negatively affects the verification by decreasing its $AA$ values. However, the impact of pruning is too high in three cases in Pong and most of the cases in MsPacman, despite failed verification. Based on our robustness definition in Section~\ref{ssec:adversarymodel}, we conclude that \fingerprint is robust against model modification attacks.    

\subsubsection{Robustness Against Evasion Attacks and Well-informed Adversaries.}\label{ssec:robustnessevasion} \adversary can evade verification by discovering individual inputs used for verification or adapt the agent's behavior to avoid a successful verification. For evasion, \adversary should have more information about the ownership verification procedure. In our setup, \adversary knows that the ownership verification is done via \fingerprint but is unaware of the exact adversarial mask used during verification. Based on this information, the simplest evasion attack is performing suboptimal actions with a pre-defined random action ratio on each episode. Figure~\ref{fig:evasion_results} confirms that the increase in the random action ratio causes a decrease in agent performance (lower return) despite successful evasion. Therefore, \fingerprint is robust against evasion via suboptimal action return.

Evasion attacks can combine detecting adversarial examples (i.e., fingerprints used for verification) and then performing either suboptimal actions or restoring original actions. We employ Visual Foresight (VF)~\cite{lin2017detecting} to carry out this attack. VF predicts the next states and the associated probability distribution of actions by looking at a history of previous states and observed actions. If the distance between the predicted and current action distribution is large, VF detects that state as adversarial, and 
performs the predicted action instead of the current one as a recovery mechanism. Figure~\ref{fig:evasion_results} shows that the use of VF does not affect the agent performance. However, the high values of $AA$ shown in the figure justifies that VF cannot recover agent performance when states are perturbed with non-transferable, universal adversarial masks and fail to evade verification. This is because the collected history of previous states consists of adversarial inputs, which might lead to the original (incorrect) action even if the adversarial state is detected correctly~\cite{tekgul2022real}. For this reason, we also evaluated the case where VF chooses a suboptimal action (VF + suboptimal action) instead of the one predicted during recovery. Figure~\ref{fig:evasion_results} shows that it decreases $AA$ more than VF, but $AA$ is not too low to evade verification and change the final verdict.     

Finally, we evaluated \fingerprint against the most well-informed adversaries that can improve the robustness of the policy against $l_{\infty}$ norm adversarial perturbations by adversarial training. For adversarial training, we implemented one of the recent state-of-the-art methods, RADIAL-RL~\cite{oikarinen2021robust}. We choose to implement RADIAL-RL for DQN agents, because these are shared cases between our and the authors' experiments. RADIAL-DQN (RADIAL-RL designed for DQN) first obtains a policy without adversarial training and then fine-tunes the policy by incorporating an adversarial loss term into the loss function that is minimized during training. In our setting, \adversary performs RADIAL-DQN by skipping the first step and fine-tunes the stolen policy $\pi_{\mathcal{A}}$ using adversarial loss. We adopted the open source repository of the authors~\footnote{\url{https://github.com/tuomaso/radial_rl_v2}} in our framework, did not change the hyperparameters used in RADIAL-DQN, and saved both agents with the best performance and the final agent after RADIAL-DQN was completed. 

The first two rows of Table~\ref{tab:radial-rl} summarize the impact, $AA$ values, and the votes for agents modified through RADIAL-DQN. The results indicate that \adversary can evade verification by making $\pi_{\mathcal{A}}$ more robust to adversarial states in Pong. \adversary obtains an improved policy for MsPacman (3rd column, negative impact: higher reward), but cannot evade verification. This outcome is not surprising, as DNN fingerprinting has limitations against adaptive adversaries that perform adversarial training~\cite{lukas2021deep}. Then, we considered an alternative scenario where \victim fine-tunes its policy with RADIAL-DQN, saves the best agent, and generates fingerprints for this agent (RDQN). The last two rows of Table~\ref{tab:radial-rl} show the verification results when \adversary implements RADIAL-DQN against adversarially robust victim agents. In this case, \adversary cannot evade verification without affecting the agent's performance. Therefore, although \fingerprint is limited against adversarial training, it satisfies the robustness requirement when fingerprinting adversarially robust victim agents.  

\begin{table}[t]
\centering
\caption{Average impact, $AA$ and voting results for stolen policies modified by RADIAL-DQN. Results are reported for both the agent with the best performance during RADIAL-DQN (3rd column) and the final agent obtained after RADIAL-DQN finishes (4th column). $AA$ is averaged on 10 verification episodes and impact is averaged over 10 test episodes. (*: improved policy, \protect\greenline : Successful verification with $AA\geq 0.75$, \protect\blueline : Successful verification with $0.75 \geq AA \geq 0.50$, \protect\yellowline : Failed verification with high impact $\geq 0.4$, \protect\redline : Failed verification with low impact $ < 0.4$)}\label{tab:radial-rl}
\begin{tblr}{colspec={cccc},
cell{5}{3-4} = {red!40},
cell{8}{3} = {blue!30},
cell{8}{4} = {red!40},
cell{11}{3-4} = {green!40},
cell{14}{3} = {blue!30},
cell{14}{4} = {yellow!40}
}
\SetCell[r=2]{m,5em}\bf Game, DRL method & \SetCell[r=2]{m}\bf Stats & \SetCell[r=2]{m}\bf Best Agent&  \SetCell[r=2]{m}\bf Final Agent \\
 & &  & \\
\hline
\SetCell[r=3]{m,5em}\bf Pong, RADIAL-DQN & Impact & $0.0 \pm  0.0$ & $0.0 \pm  0.0$\\
                                        & $AA$ & $0.04 \pm 0.06$ & $0.04 \pm 0.06$\\ 
                                        & Votes &  0 \cmark / 10 \xmark &  0 \cmark / 10 \xmark\\
\hline[dashed]
\SetCell[r=3]{m,5em}\bf MsPacman, RADIAL-DQN & Impact & $-0.16 \pm  0.03^{*}$  & $ 0.39 \pm 0.03$\\
                                        & $AA$ &  $0.59 \pm 0.40$ &  $0.29 \pm 0.31$\\ 
                                        & Votes & 6 \cmark / 4 \xmark & 4 \cmark / 6 \xmark\\
\hline[dashed]
\SetCell[r=3]{m,5em}\bf Pong, RADIAL-RDQN & Impact & $0.0 \pm 0.0$ & $0.0 \pm  0.0$\\
                                        & $AA$ &$0.84 \pm 0.21$ & $0.89 \pm 0.17$\\ 
                                        & Votes & 8 \cmark / 2 \xmark & 9 \cmark / 1 \xmark  \\
\hline[dashed]
\SetCell[r=3]{m,5em}\bf MsPacman, RADIAL-RDQN & Impact & $0.15 \pm 0.04$ & $0.55 \pm 0.06$\\
                                        & $AA$ & $0.61 \pm 0.34$& $0.09 \pm 0.18$\\ 
                                        & Votes & 7 \cmark / 3 \xmark & 1 \cmark / 9 \xmark \\
\hline
\end{tblr}
\end{table}

\begin{table*}[t]
\centering
\caption{$AA$ values (averaged over 10 verification episodes) and voting results for false claims against victim \victim, and independent $\mathcal{I}$ policies with different perturbation constraint $\epsilon$ values. (The cases where a false claim succeeds are shown as follows: \protect\greenline : False claim with $AA\geq 0.75$, \protect\blueline : False claim with $0.75 \geq AA \geq 0.50$)}\label{tab:moc}
\begin{tblr}{colspec={cccccc},
cell{7}{3-4} = {blue!30},
cell{7}{5-6} = {green!40},
cell{8}{3-6} = {blue!30},
cell{14}{4-5}= {blue!30},
cell{14}{6}  = {green!40},
}
 &  & \SetCell[c=4]{c}\bf $\bm{\epsilon}$ vs. $\bm{AA}$ (Votes) & & &\\
\cline{3-6}
\SetCell[c=2]{c} \bf Game, DRL method & & \bf 0.05 & \bf 0.1 & \bf 0.2 & \bf 0.5 \\
\hline
\bf Pong, & \victim & $0.45\pm 0.47$ (5 \cmark / 5 \xmark)& $0.49\pm 0.49$ (5 \cmark / 5 \xmark) & 
$0.40\pm 0.49$ (4 \cmark / 6 \xmark)& $0.40\pm 0.49$ (4 \cmark / 6 \xmark)\\
\bf A2C & $\bm{\mathcal{I}}$, avg. & $0.32 \pm 0.36$ (3 \cmark / 7 \xmark)& $0.38\pm 0.45$ (3 \cmark / 7 \xmark) & $0.30 \pm 0.41$ (3 \cmark / 7 \xmark) & $0.28 \pm 0.43$ (3 \cmark / 7 \xmark)\\ 
\hline[dashed]
\bf Pong, & \victim & $0.37 \pm 0.42$ (4 \cmark / 6 \xmark)& $0.37 \pm 0.45$ (3 \cmark / 7 \xmark) & 
$0.33 \pm 0.45$ (3 \cmark / 7 \xmark)  & $0.40 \pm 0.49$ (4 \cmark / 6 \xmark)  \\
\bf DQN &  $\bm{\mathcal{I}}$, avg.& $0.01 \pm 0.18$ (1 \cmark / 9 \xmark)& $0.07 \pm 0.22$ (1 \cmark / 9 \xmark)& $0.05 \pm 0.19$ (1 \cmark / 9 \xmark) & $0.05 \pm 0.19$ (1 \cmark / 9 \xmark)  \\ 
\hline[dashed]
\bf Pong, & \victim &  $0.56 \pm 0.39$ (5 \cmark / 5 \xmark)& $0.68 \pm 0.42$ (7 \cmark / 3 \xmark)& $0.76 \pm 0.38$ (8 \cmark / 2 \xmark) & $0.78 \pm 0.39$ (8 \cmark / 2 \xmark)  \\
\bf PPO &  $\bm{\mathcal{I}}$, avg. & $0.56 \pm 0.36$ (6 \cmark / 4 \xmark)& $0.59 \pm 0.38$ (6 \cmark / 4 \xmark)& $0.59 \pm 0.38$ (6 \cmark / 4 \xmark) & $0.52 \pm 0.41$ (6 \cmark / 4 \xmark) \\ 
\hline[dashed]
\bf MsPacman, & \victim & $0.00\pm 0.00$ (0 \cmark /10 \xmark) & $0.03\pm 0.05$ (0 \cmark /10 \xmark) &  $0.14\pm 0.29$ (1 \cmark /9 \xmark)  & $0.09\pm 0.22$ (1 \cmark /9 \xmark)  \\
\bf A2C &  $\bm{\mathcal{I}}$, avg.& $0.15\pm 0.56$  (1 \cmark /9 \xmark)  & $0.14\pm 0.21$ (1 \cmark /9 \xmark)  & $0.13\pm 0.30$ (2 \cmark /8 \xmark) & $0.21\pm 0.36$ (2 \cmark /8 \xmark) \\ 
\hline[dashed]
\bf MsPacman, & \victim & $0.23\pm 0.36$ (2 \cmark /8 \xmark)& $0.0\pm 0.0$ (0 \cmark /10 \xmark) & $0.0\pm 0.0$ (0 \cmark /10 \xmark) & $0.0\pm 0.0$ (0 \cmark /10 \xmark) \\
\bf DQN &  $\bm{\mathcal{I}}$, avg.& $0.26\pm 0.24$ (2\cmark /8 \xmark)  & $0.19\pm 0.26$ (2\cmark /8 \xmark)& $0.15 \pm 0.29$ (1\cmark /9\xmark) & $0.24 \pm 0.26$ (3\cmark /7\xmark)\\ 
\hline[dashed]
\bf MsPacman, & \victim & $0.19 \pm 0.18$ (1 \cmark / 9 \xmark) & $0.26 \pm 0.31$ (3 \cmark / 7 \xmark) & $0.38 \pm 0.37$ (4 \cmark / 6 \xmark) & $0.07 \pm 0.21$ (1 \cmark / 9 \xmark)  \\
\bf PPO &  $\bm{\mathcal{I}}$, avg.& $0.10\pm 0.11$  (0 \cmark /10 \xmark) & $0.50\pm 0.39$ (5 \cmark /5 \xmark) & $0.74\pm 0.40$ (8 \cmark /2 \xmark) & $0.80\pm 0.20$ (8 \cmark /2 \xmark)\\ 
\hline[dashed]
\end{tblr}
\end{table*}

\subsubsection{Robustness Against False Claims.}\label{ssec:robustnessfalse} 
Liu et al.~\cite{liu2023false} show that malicious accusers can produce fake fingerprints that can pass the ownership verification test against independent models in many ownership verification schemes, including CAE~\cite{lukas2021deep}. Therefore, we also evaluated the robustness of \fingerprint against malicious accusers by generating fingerprints for the accuser's policy without maximizing the loss for independent policies (Equation~\ref{eq:lossfunction}) and not measuring the non-transferability score (Algorithm~\ref{alg:fingerprint_generation}, line 6), which is similar to the setup proposed in~\cite{liu2023false} to evaluate CAE. We selected one of the five independent policies $\pi_{i}, i\in \mathcal{I}$ that behaves the closest to $\pi_{\mathcal{V}}$ in the test episodes and has the same DRL algorithm as the accuser policy and other independent policies as \verifier's control set. As shown in Table~\ref{tab:moc}, the malicious accuser cannot falsely claim ownership of \victim for the perturbation constraint set in \fingerprint ($\epsilon=0.05$), except the PPO agent trained for Pong. If the perturbation constraint becomes larger ($\epsilon \geq 0.1$), then the accuser's false fingerprints transfer to other models in those cases. Having \verifier perform an additional check that the size of the adversarial mask does not exceed a prescribed bound can mitigate against false claims attacks for \fingerprint. Table~\ref{tab:moc} also indicates that adversarial states have a higher transferability rate between PPO algorithms compared to others. In these cases, \verifier can train or search for other independent PPO policies for the same task as suggested in~\cite{liu2023false}, and it can reject the claim if the accuser's fingerprints falsely verify all independent models. Therefore, we conclude that \fingerprint is not susceptible to false claims with a simple additional countermeasure on $\epsilon$ and non-transferability check based on the DRL algorithm. 

\textit{Model extraction attacks in DRL : } In this work, we limit the scope to the adversary model described in Section~\ref{ssec:adversarymodel} and do not consider model extraction attacks against DRL policies through imitation learning~\cite{chen2021stealing}. Nevertheless, we tried to implement the model extraction attack proposed by Chen et al.~\cite{chen2021stealing}, but were unable to obtain good stolen policies, which could be due to the simpler tasks chosen in the setup of the original work. Chen et al.~\cite{chen2021stealing} experimentally show that adversarial examples can successfully transfer from stolen policies to the victim policy if they share the same DRL algorithm. Their preliminary results provide insight into the possibility of preserving fingerprints during the extraction of the DRL model. Thus, we leave the construction of effective DRL model extraction attacks and evaluate the robustness of \fingerprint against these attacks for future work.

\section{Transferability of Universal Masks}
\label{sec:Discussion}

The fingerprint generation process in \fingerprint is based on maximum-confidence adversarial example generation techniques and is similar to Fast Gradient Sign Method (FGSM)~\cite{huang2017adversarial}, since \fingerprint averages the gradient of Equation~\ref{eq:lossfunction} w.r.t. randomly selected states. As presented in Section~\ref{ssec:adversarial}, maximum-confidence adversarial examples have a higher transferability rate than minimum-confidence examples. FGSM is a maximum-confidence method itself; however, during the computation of universal, non-transferable adversarial masks, the effect of the high-sensitivity directions obtained from the most confident adversarial examples is diminished by others. Nevertheless, we analyzed whether minimum-confidence adversarial masks in DNNs can be useful for DRL fingerprinting. 
For that reason, we changed the universal mask generation $\vect{r}$, (Algorithm~\ref{alg:fingerprint_generation}, line 4) with Universal Adversarial Perturbation (UAP)~\cite{moosavi2017universal} by implementing the method proposed for DRL settings~\cite{tekgul2022real}.

\begin{figure}[t]
\centering
\includegraphics[width=1.0\columnwidth]{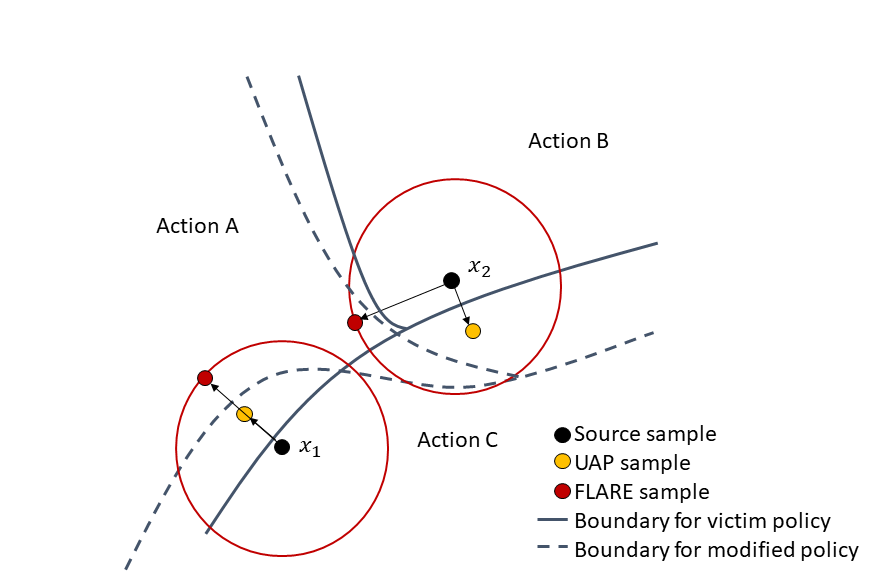} 
\caption{Depiction of fingerprints generated by UAP and \fingerprint for a DRL policy and three available actions. UAP moves the source sample in the direction of the closest incorrect action, and typically this movement is less than the perturbation constraint (denoted by circles). In contrast, \fingerprint shifts the source sample to the same action, which is irrelevant to the original action, while using the maximum value of the perturbation constraint.}
\label{fig:discussion}
\end{figure}

\begin{table*}[t]
\centering
\caption{Comparison of UAP and \fingerprint based on fooling rate (measured for the victim policy) and action agreement $AA$. Both the fooling rate and $AA$ are averaged using adversarial states (fingerprints) in 10 verification episodes. The higher fooling rate and $AA$ values are highlighted in green. Matched actions: Cases where victim and modified policies perform the same action for the same state. Different actions: Cases where victim and modified policies perform different actions for the same state.}\label{tab:uap}
\begin{tabular}{c |cccc |cccc}
\multicolumn{1}{c}{\bf Game, DRL Method} & \multicolumn{4}{c}{\bf UAP} & \multicolumn{4}{c}{\bf \fingerprint} \\
\hline
\bf (Fine-tuned & \multicolumn{2}{c}{\bf Matched actions} &  \multicolumn{2}{c|}{\bf Different actions} & \multicolumn{2}{c}{\bf Matched actions} &  \multicolumn{2}{c}{\bf Different actions} \\
\bf over 200 eps.) & Fooling rate &  $AA$ & Fooling rate &  $AA$ & Fooling rate &  $AA$ & Fooling rate &  $AA$ \\
\hline
\bf Pong, A2C  & $0.79 \pm 0.09 $& $0.78 \pm 0.08$&  $\cellcolor{green!40}{0.89 \pm 0.07}$& $0.69 \pm 0.13$& $\cellcolor{green!40}{0.95 \pm 0.10}$& $\cellcolor{green!40}{0.94 \pm 0.09}$ & $0.85\pm 0.12$& $\cellcolor{green!40}{0.92 \pm 0.16}$\\
\bf Pong, DQN  & $0.69 \pm 0.11 $& $0.12 \pm 0.10$& $0.55\pm 0.18$& $0.24 \pm 0.14$& 
$\cellcolor{green!40}{0.89 \pm 0.13}$& $\cellcolor{green!40}{0.92 \pm 0.18}$& $\cellcolor{green!40}{0.93 \pm 0.07}$& $\cellcolor{green!40}{0.94\pm 0.09}$\\
\bf Pong, PPO  & $0.86 \pm 0.05$& $0.40 \pm 0.21$& $0.82 \pm 0.14$& $0.41 \pm 0.13$& 
 $\cellcolor{green!40}{0.91\pm 0.03}$& $\cellcolor{green!40}{0.98 \pm 0.08}$ & $\cellcolor{green!40}{0.90 \pm 0.07}$ & $\cellcolor{green!40}{0.87 \pm 0.3}$\\
\bf MsPacman, A2C  &  $\cellcolor{green!40}{0.76 \pm 0.32}$ & $0.53 \pm 0.42 $& $\cellcolor{green!40}{0.91 \pm 0.13}$& $\cellcolor{green!40}{0.58 \pm 0.42}$& $0.68 \pm 0.36$& $\cellcolor{green!40}{0.64 \pm 0.36}$& $0.64 \pm 0.42$ &$0.55 \pm 0.43$\\
\hline
\end{tabular}
\end{table*}

We found that \fingerprint with UAP satisfies the effectiveness and integrity requirements for all agents, except the DQN agent trained for MsPacman. It was impossible to obtain an adversarial example with the perturbation constraint used in \fingerprint ($\epsilon = 0.05$) against this agent, but increasing it leads to transferable adversarial examples and false positives. The real issue with UAP emerges when the adversary \adversary modifies the stolen policy $\pi_{\mathcal{A}}$ with model modification attacks. Due to its minimum-distance property, UAP finds the smallest high-sensitivity directions belonging to the closest incorrect class (or discrete actions in DRL), and generally the resulting $\vect{r}$ is smaller than $\epsilon$. Therefore, a small change in $\pi_{\mathcal{A}}$ negatively affects the robustness of UAP. Contrary to UAP, \fingerprint shifts the source sample using the maximum amount of perturbation $\epsilon$, forces $\pi_{\mathcal{A}}$ to perform the same incorrect action and is more robust against model modification attacks. We illustrate this problem in Figure~\ref{fig:discussion}.   

One of the main reasons why \fingerprint has better robustness stems from the fact that the input space embeddings in DRL are not as separable as in DNN~\cite{annasamy2019towards}. In DRL, although the input states are spatially similar, they often result in different actions. 
DRL agents optimize policies using both input state and environment dynamics and act upon spatio-temporal abstractions~\cite{zahavy2016graying}. \fingerprint identifies discontinuities in the optimal policy and computes an adversarial state that is spatially similar to the source state but far from it in temporal dimension. 
UAP typically explores adversarial pockets that are closer in the spatial domain due to its minimum-distance strategy. Therefore, it cannot withstand model modification attacks that preserve the spatio-temporal abstractions and slightly change the sequential strategy. 

We provide experimental results for our discussion in Table~\ref{tab:uap}. This table compares the fooling rate and action agreement $AA$ for adversarial states used in the verification of fine-tuned policies. We chose to report the results for fine-tuned policies from Table~\ref{tab:fine_tuning_pruning} considering the acceptable impact range $(< 0.4)$ on the modified agent's performance. Matched actions refers to situations where both victim and modified policies perform the same action for the same input state without any added fingerprints. In contrast, different actions refer to cases where victim and modified policies behave differently for the same input state. Table~\ref{tab:uap} shows that the fooling rate of UAP is lower than \fingerprint in almost all cases. This supports our first claim regarding the robustness of UAP and \fingerprint. The columns labeled with $AA$ show the action agreement where the fingerprint successfully misleads the victim policy. In this case, the ideal $AA$ value for matched actions would be $1.0$. As can be seen from Table~\ref{tab:uap}, \fingerprint reaches much higher $AA$ values for matched actions than UAP. Surprisingly, \fingerprint attains higher $AA$ values for different actions as well. This shows that, even if the fine-tuned policy successfully changes the agent's behavior, the adversarial states generated by \fingerprint force the policy to perform the same incorrect action. The same conclusion cannot be drawn from the UAP results, as the $AA$ values reported for the same actions are lower than \fingerprint even in the case with the higher fooling rate.

Based on this discussion, we conjecture that using minimum-distance adversarial examples to fingerprint DRL agents requires either adding modified policies to the loss function in Equation~\ref{eq:lossfunction}, or considering the temporal structure of the policy while finding the high-sensitivity directions. The latter option also opens a new space of adversarial examples that can exploit temporal abstractions learned by DRL policies.  

\section{Related Work}
\label{sec:Related}

\textit{Adversarial Examples in DRL :} Recent work has shown that DRL policies are vulnerable to adversarial examples generated for agents' states~\cite{huang2017adversarial,tekgul2022real} or actions~\cite{weng2020toward} in single-agent environments, or produce natural adversarial states by exploiting other agents in multi-agent settings~\cite{gleave2019adversarial}. 
Other studies focus on perturbing the dynamics of the environment by modifying the environment conditions~\cite{MankowitzL2020Robust,pan2022characterizing}. DRL adversarial training ~\cite{oikarinen2021robust, zhang2018protecting} has been considered as a mitigation, but adversarially robust policies were found to be more vulnerable to high-sensitivity directions caused by a natural change in the environment~\cite{korkmaz2022deep}.    

\noindent\textit{Ownership Verification via Model Watermarking :} Model watermarking has become a widely known ownership verification procedure for DNNs~\cite{adi2018turning,zhang2018protecting,lounici2021yes}. Model watermarking embeds traceable information (i.e. watermark) into the DNN by either directly inserting it into model parameters or adding unique knowledge into a small subset of the training set. During ownership verification, the existence of the watermark is proven on illegitimate copies. 
Previous DRL ownership verification methods adapt model watermarking techniques. For example, Behzadan et al.~\cite{behzadan2019sequential} propose the embedding of sequential states that are separate from the main environment as watermarks during training. However, watermark verification also requires a different environment, and there is no guarantee that watermarks will be retained while learning complex tasks. Chen et al.~\cite{chen2021temporal} obtain a sequence of damage-free states as watermarks that are sampled from the same environment and do not impact agent performance. During verification, the authors compare the action probability distributions given by both the victim and the suspected agents over these sequential states. However, this watermarking method requires modifying both the training process and the reward function. 

Although model watermarking is considered a practical solution to protect DNN ownership, many studies have shown~\cite{lukas2022sok,yan2022cracking} that they cannot withstand well-informed adversaries and model modification attacks. Compared to watermarking, DNN fingerprinting methods show improved robustness to model modification and extraction attacks~\cite{lukas2021deep,peng2022fingerprinting}. Furthermore, fingerprinting does not change the training procedure unlike watermarking. However, there is no prior work applying fingerprinting as an ownership verification method in DRL.
                                
\noindent\textit{\newtext{Ownership Verification of Large models via 
Fingerprinting: }} \newtext{Although \fingerprint is specifically designed for DRL, we conjecture that universal and non-transferable adversarial masks can be useful for fingerprinting, e.g., large language models. For example, Wallace et al.~\cite{wallace2019universal} show the availability of context-independent universal adversarial triggers that force large language models to produce incorrect results. Similarly, Gu et al.~\cite{gu2022vision} demonstrate that universal adversarial patches can fool vision transformers. If universality is restricted with the non-transferability requirement, then the generated adversarial masks will profile the global behavior of large models and can be used in ownership verification.}
\section{Conclusion}
\label{sec:Conclusion}

In this paper, we propose \fingerprint, the first fingerprint method that can be used for ownership verification of DRL policies, and show the existence of non-transferable universal adversarial masks in DRL settings. We empirically demonstrate that our fingerprints are efficient and do not accidentally accuse independently trained models. Adversarial training is the only method that evades verification by making policies robust to adversarial examples. However, our experiments show that the fingerprints obtained by \fingerprint for robust policies are persistent. We hypothesize that \fingerprint can be extended to continuous tasks, where the verifier can check how much the suspected agent deviates from the original action value, and we leave this for future work.

A promising direction for future work related to DRL fingerprinting is to study whether an intentional change in environment conditions can be useful candidates for fingerprints. DRL policies show decreased robustness when deployed in a different environment and include high-sensitivity directions due to natural causes. This vulnerability leads to model evasion attacks via natural adversarial examples, but can also be leveraged to learn natural (and non-transferable) fingerprints for ownership verification. We believe that our study can create more interest in securing DRL agents using novel ownership verification methods against possible model piracy and extraction attacks.   

\subsubsection*{Acknowledgements} \newtext{This research was partially supported by Intel. We thank Dr. Samuel Marchal and Shelly Wang for initial discussions on this problem and for collaborating on an alternative approach to fingerprinting DRLs that we explored prior to the solution presented in this paper. We also thank Aalto Science-IT for computational resources.}

\bibliographystyle{ACM-Reference-Format}
\bibliography{references}

\appendix
\section{Appendix}
\label{sec:appendixA}
\subsection{Training DRL Agents}
\label{ssec:appendixA1}

\begin{table*}
\centering
\caption{Return (averaged over 10 test episodes) of the victim and independent policies trained for MsPacman. The best and worst agents for the same DRL algorithm are highlighted in green and red, respectively.}\label{tab:mspacman}
\begin{tabular}{c|c|ccccc}
\multicolumn{1}{c}{\bf DRL Method} & \multicolumn{1}{c}{\bf Victim Agent}  & \multicolumn{5}{c}{\bf Independent Agents} \\
\hline
A2C & \textcolor{green}{$3316.00 \pm  512.72$} & \textcolor{red}{$1670.00 \pm 537.27$} & $2552.00 \pm 595.66$ & $2144.00 \pm 816.58$ & $2246.00 \pm 4.90$ &  $1750.00 \pm 72.66$ 
\\
DQN & \textcolor{green}{$2620.00 \pm 80.62$} & $2363.00 \pm 269.26$ & $2484.00 \pm 389.67$ & $2218.00 \pm 347.84$ & \textcolor{red}{$2211.00 \pm 154.24$} & $2472.00 \pm 412.74$ 
\\
PPO & \textcolor{green}{$2731.00 \pm 545.50$} & $ 2019.00 \pm 77.13$ &  $2198.00\pm536.35$ & $2040.00 \pm 161.43$ & \textcolor{red}{$2017.00 \pm 397.57$} & $2167.00 \pm 268.52$ 
\\
\hline
\end{tabular}
\end{table*}

To facilitate the comparison, we used the same setup to implement all DRL policies and attacks: PyTorch (version 1.4.0), NumPy (version 1.18.1), Gym (a toolkit for developing reinforcement learning algorithms, version 0.15.7) and Atari-Py (a Python interface for the Arcade Learning Environment,
version 0.2.6). All experiments were carried out on a computer with 2x12 core Intel(R) Xeon(R) CPUs (32GB RAM) and NVIDIA Quadro P5000 with 16GB memory. To train DQN agents, we used a dueling Q-network architecture proposed in~\cite{wang2016dueling}. For training A2C and PPO agents, we choosed to implement the convolutional neural networks suggested in OpenAI Baselines\footnote{\url{https://github.com/openai/baselines}}. In both A2C and PPO, actors and critics use the same architecture, except for the penultimate later. The hyperparameter values of each victim agent are set the same as OpenAI baselines, while slightly differ for training independent agents.

All victim and independent DRL agents trained to play Pong reach the highest score +21. The summary of agents trained to play MsPacman is presented in Table~\ref{tab:mspacman}. In MsPacman, we deliberately chose agents with the best performance as the victim, since they have a clear business advantage over other models, thus incentivizing adversaries to apply piracy attacks against them.

\begin{table}
\centering
\caption{Hyperparameters used in fingerprint generation}\label{tab:hyperparameters}
\resizebox{1.0\columnwidth}{!}{%
\begin{tblr}{ccc}
\bf Parameter & \bf Value  & \bf Definition \\
\hline
$\epsilon$ & $0.05$ & $l_{\infty}$ constraint on the perturbation $\vect{r}$ \\
$\tau_{nts}$ & $0.7$& minimum non-transferability score of $\vect{r}$\\
$\tau_{\delta}$ & $0.8$& minimum fooling rate of $\vect{r}$ on a dataset\\
\SetCell[r=2]{m} $n_{\text{episodes}}$ & \SetCell[r=2]{m} $1000$ & maximum number of training episodes \\
& & to generate/collect fingerprints \\
\hline
\end{tblr}%
}
\end{table}

\subsection{Hyperparameter Selection in \fingerprint}
\label{ssec:appendixA3}

The perturbation constraint $\epsilon$ directly affects the trade-off between the success of an adversarial example and its non-transferability. Therefore, we performed a grid search for $\epsilon$ and set it to an optimal value $0.05$. We set the minimum fooling rate $\tau_{\delta}$ at a high value $0.8$ to ensure the universality of the adversarial mask and set the non-transferability score at $0.7$. Based on these values and Equation~\ref{eq:nts}, for a candidate universal adversarial mask $\vect{r}$, the minimum action agreement of independent agents $ min_{i \in \mathcal{I}}(AA(\pi_{\mathcal{V}}, \pi_{i}, \vect{s}, \vect{r}))$ should be lower than $0.125$ to be chosen as a valid fingerprint. For both the Pong and MsPacman agents, we used a reduced set of actions (4 discrete actions in total). The minimum $ min_{i \in \mathcal{I}}(AA(\pi_{\mathcal{V}}, \pi_{i}, \vect{s}, \vect{r})) = 0.125$ is much lower than $0.25$ ($AA$, if actions are randomly chosen) and satisfies the non-transferability requirement. Finally, we set $n_{\text{episodes}}$ at a high value to guarantee that a sufficient number of fingerprints is generated for efficient verification. The prescribed hyperparameter values during fingerprint generation are listed in Table~\ref{tab:hyperparameters}.  

\subsubsection{Selection of the Number of Fingerprints}

\begin{figure}
\centering
\includegraphics[width=1.0\linewidth]{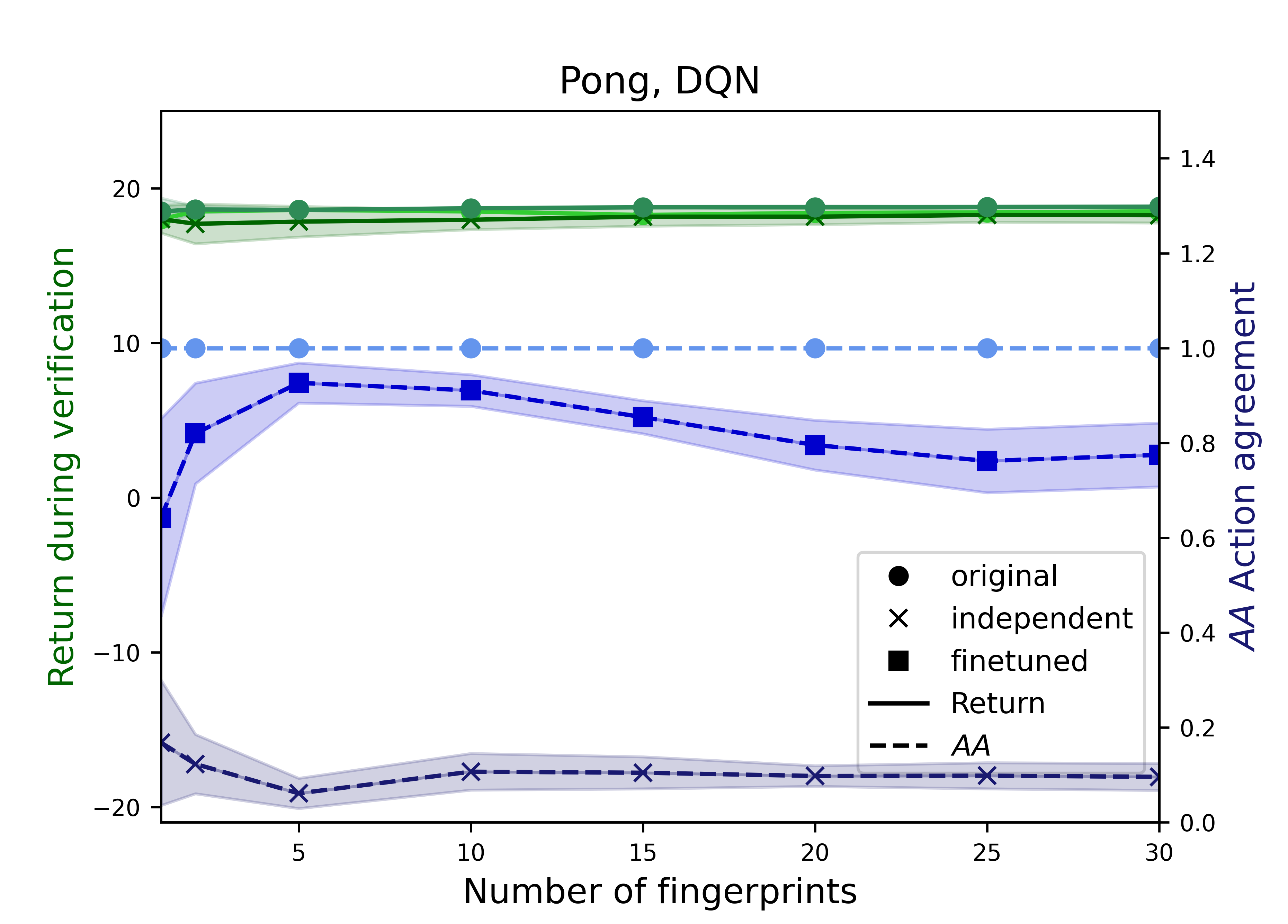}
\caption{The effect of the number of fingerprints on the return during verification and $AA$ averaged over 10 verification episodes.}\label{fig:fingerprints_onPong}
\end{figure}

The number of fingerprints generated and used for verification affects integrity and robustness. An insufficient number of fingerprints could result in a high action agreement $AA$ between the independent and victim (original) policies and ultimately falsely verify the ownership of the independent policies as explained in Section~\ref{ssec:reliability}. On the contrary, a high number of fingerprints could give low $AA$ between the victim policy and its modified versions, since some of the fingerprints could give a lower fooling rate in the modified policies. For that, we performed verification by changing the maximum number of fingerprints used for fingerprint generation. As demonstrated in Figure~\ref{fig:fingerprints_onPong}, the number of fingerprints does not affect the return during verification, but a sufficient number of fingerprints (around 5) are needed to achieve high $AA$ to provide high confidence for the final decision. We set the number of fingerprints at $10$ to satisfy the effectiveness and robustness requirements simultaneously. 

\subsubsection{Selection of the Window Size.} 
\begin{figure}
\centering
\includegraphics[width=1.0\linewidth]{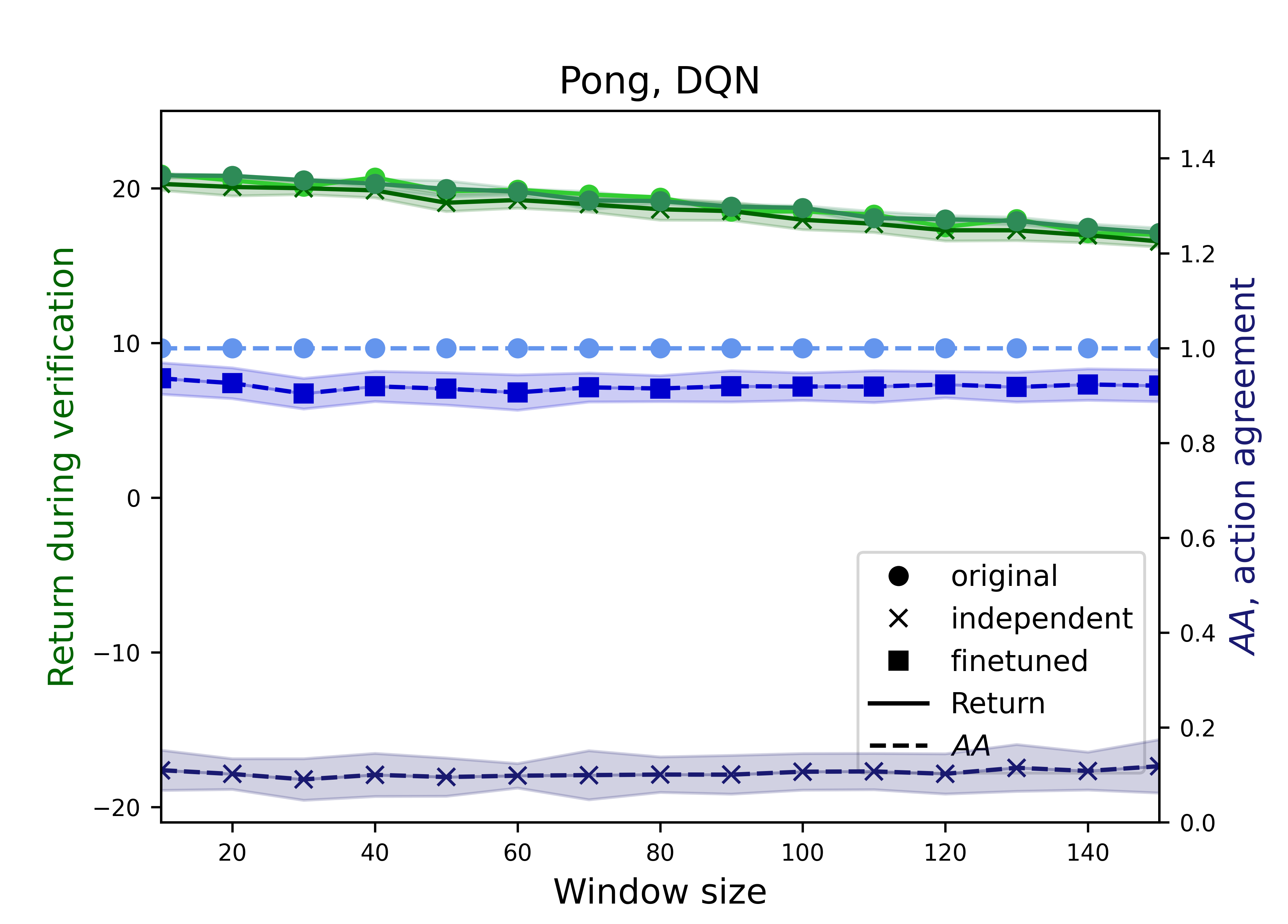}
\caption{The effect of window size on the return during verification and $AA$ averaged over 10 verification episodes.}\label{fig:windowsize_onPong}
\end{figure}

In addition to the number of fingerprints, the decision on window size is important. If the window size is large, then the return during verification decreases and the agent can perform poorly. Figure~\ref{fig:windowsize_onPong} illustrates the effect of window size on return and $AA$ during verification for Pong DQN agents. Although $AA$ does not change significantly with larger window sizes, there is a steady decline in return. Based on this result, the window size can be set to $40$ or even less, but we set it to $40$ after performing the same analysis for all agents and observing the change in return.   

\subsubsection{Computation costs of \fingerprint}

\begin{table}
\centering
\caption{\newtext{Total number of trials (i.e., episodes) required for obtaining the fingerprint list during the fingerprint generation phase (2nd column), and the average ratio of adversarial states that includes the fingerprint to the total number of states observed for the same episode (3rd column) during verification. The ratio is averaged over 10 verification episodes for each victim agent.}}\label{tab:costoffingrprint}
\resizebox{1.0\columnwidth}{!}{%
\begin{tabular}{ccc}
 & \bf $\#$ of trials & \bf ($\#$ of adversarial states)/($\#$ of states) \\
 & \bf in generation & \bf in verification \\ 
\hline
Pong, A2C & $34$ & $0.02 \pm 0.00$\\
Pong, DQN & $46$ & $0.02 \pm 0.00$\\
Pong, PPO & $14$ & $0.02 \pm 0.00$\\
MsPacman, A2C & $110$ & $0.04 \pm 0.01$\\
MsPacman, DQN & $38$ &  $0.05 \pm 0.01$\\
MsPacman, PPO & $10$ &  $0.05 \pm 0.01$\\
\hline
\end{tabular}%
}
\end{table}

\newtext{Based on the hyperparameters chosen in our experimental setup, we computed the number of trials (i.e. epiosodes) required to generate the fingerprint list and presented them in Table~\ref{tab:costoffingrprint}. The required number of trials is less than 50 in almost all cases, except MsPacman. We found that this exception occurs due to the high $AA$ generated for some independent policies used during the fingerprint generation phase. To generate each $\vect{r}_{candidate}$ (see Algorithm~\ref{alg:fingerprint_generation}, line 4), \fingerprint randomly selects 100 states and computes the average gradient using those states. During verification, based on the window size ($40$) and the number of fingerprints ($10$), the suspected models are queried $400$ times in total with the additional fingerprint. Table~\ref{tab:costoffingrprint} also shows the average ratio of states with an additional fingerprint to the total number of states observed during verification episodes. Based on these results, we confirm that verification episodes include only a small number of states (up to $5\%$) with the additional fingerprint.}

\subsection{Receiver Operating Characteristic of \fingerprint}
\label{ssec:appendixA4}

\begin{figure}
\centering
\includegraphics[width=1.0\linewidth]{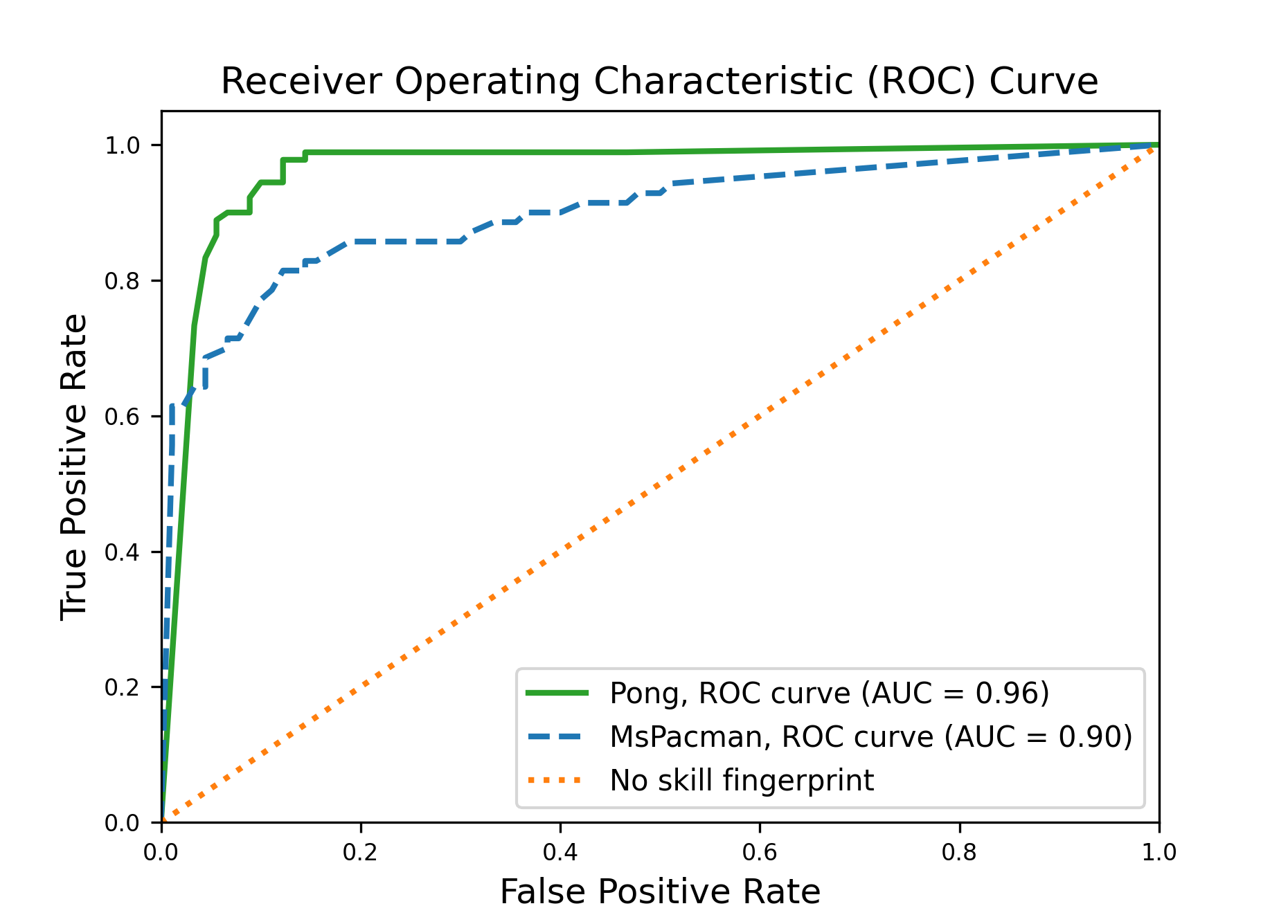}
\caption{\newtext{The effect of the threshold for individual fingerprint's decision on the verification. Results are computed over 10 fingerprints used for the verification of victim policies, randomly selected independent policies and fine-tuned policies.}}\label{fig:roc_curve}
\end{figure}

\newtext{Figure~\ref{fig:roc_curve} shows the receiver operation characteristic (ROC) curve produced by the verification results of individual fingerprints over multiple thresholds $\tau_{AA}$, where the $i$-th fingerprint votes ``stolen'' when $AA_{i} \geq \tau_{AA}$. For each victim policy, we calculated true positive and false positive rates (TPR and FPR) on 10 fingerprints that are used to verify the victim policy itself, 3 randomly selected independent policies, and 3 fine-tuned versions of the victim policy incurring a small impact on utility. We found the optimal $\tau_{AA}$ that maximizes TPR and minimizes FPR to be $0.5$ and $0.68$ in Pong and MsPacman, respectively. We set the threshold value at $0.5$ in all our experiments, but it would be beneficial to analyze the ROC for each environment, as the choice of $\tau_{AA}$ affects the overall effectiveness and integrity of \fingerprint.} 

\subsection{Impact of Verification}
\label{ssec:appendixA5}

As discussed in Section~\ref{ssec:adversarymodel}, utility is not a necessary requirement in fingerprinting methods, as fingerprints typically trigger abnormal behavior. However, the impact on agent performance is still important, since verification might also be carried out in a stealthy way to avoid raising any suspicion. Moreover, if the agent fails to perform the task quickly during verification, then the collected information may not be sufficient to correctly calculate the action agreement $AA$. Therefore, we computed the impact of verification on agent performance and summarized the results in Table~\ref{tab:vimpact}. In Pong, the impact is almost zero, while we experienced an average impact of $0.22$ in MsPacman agents due to the high complexity of the game. The impact in MsPacman can be further improved by adding fingerprints in non-critical states that do not affect the return if the agent replaces one action with another.

\begin{table}
\centering
\caption{Impact of verification (averaged over 10 verification episodes) on victim agent performance.}\label{tab:vimpact}
\begin{tabular}{c|c|cc}
\multicolumn{1}{c}{\bf DRL Method} & \multicolumn{1}{c}{\bf Pong}  & \multicolumn{1}{c}{\bf MsPacman} \\
\hline
A2C & $0.02 \pm 0.02$ & $0.20 \pm 0.21$ \\
DQN & $0.01 \pm 0.02$ & $0.28 \pm 0.19$ \\
PPO & $0.02 \pm 0.02$ & $0.18 \pm 0.28$ \\
\hline
\end{tabular}
\end{table}

\end{document}